\documentclass[10pt,twocolumn,letterpaper]{article}

\usepackage[pagenumbers]{cvpr} 

\usepackage[dvipsnames]{xcolor}

\usepackage{array}
\usepackage{graphicx}
\usepackage{amsmath}
\usepackage{amssymb}
\usepackage{booktabs}

\usepackage{booktabs}
\usepackage{multirow}
\usepackage{multicol}
\usepackage{makecell}
\usepackage{caption,subcaption}
\usepackage{marvosym}
\usepackage[section]{placeins}

\newcommand\blfootnote[1]{%
  \begingroup
  \renewcommand\thefootnote{}\footnote{#1}%
  \addtocounter{footnote}{-1}%
  \endgroup
}
\makeatletter
\newcommand*\bigcdot{\mathpalette\bigcdot@{.5}}
\newcommand*\bigcdot@[2]{\mathbin{\vcenter{\hbox{\scalebox{#2}{$\m@th#1\bullet$}}}}}
\makeatother

\definecolor{cvprblue}{rgb}{0.21,0.49,0.74}
\usepackage[pagebackref,breaklinks,colorlinks,citecolor=cvprblue]{hyperref}

\usepackage[capitalize]{cleveref}
\crefname{section}{Sec.}{Secs.}
\Crefname{section}{Section}{Sections}
\Crefname{table}{Table}{Tables}
\crefname{table}{Tab.}{Tabs.}

\def\paperID{3499} 
\def\confName{CVPR}
\def\confYear{2024}

\title{ART$\bigcdot$V: Auto-Regressive Text-to-Video Generation with Diffusion Models \\
}

\author{Wenming Weng\textsuperscript{1,2~$\ast$}, Ruoyu Feng\textsuperscript{1,2~$\ast$}, Yanhui Wang\textsuperscript{1,2~$\ast$}, Qi Dai\textsuperscript{2}, Chunyu Wang\textsuperscript{2}, Dacheng Yin\textsuperscript{1,2~$\ast$}, \\
Zhiyuan Zhao\textsuperscript{2},
Kai Qiu\textsuperscript{2}, Jianmin Bao\textsuperscript{2}, Yuhui Yuan\textsuperscript{2}, Chong Luo\textsuperscript{2~$\dagger$}, Yueyi Zhang\textsuperscript{1}, Zhiwei Xiong\textsuperscript{1}\\
{\textsuperscript{1}University of Science and Technology of China
~~\textsuperscript{2}Microsoft Research Asia}\\
{\url{https://warranweng.github.io/art.v}}
}

\begin{document}
\twocolumn[{
\renewcommand\twocolumn[1][]{#1}
\maketitle

\begin{center}
    \vspace{-5mm}
    \includegraphics[width=1.0\linewidth]{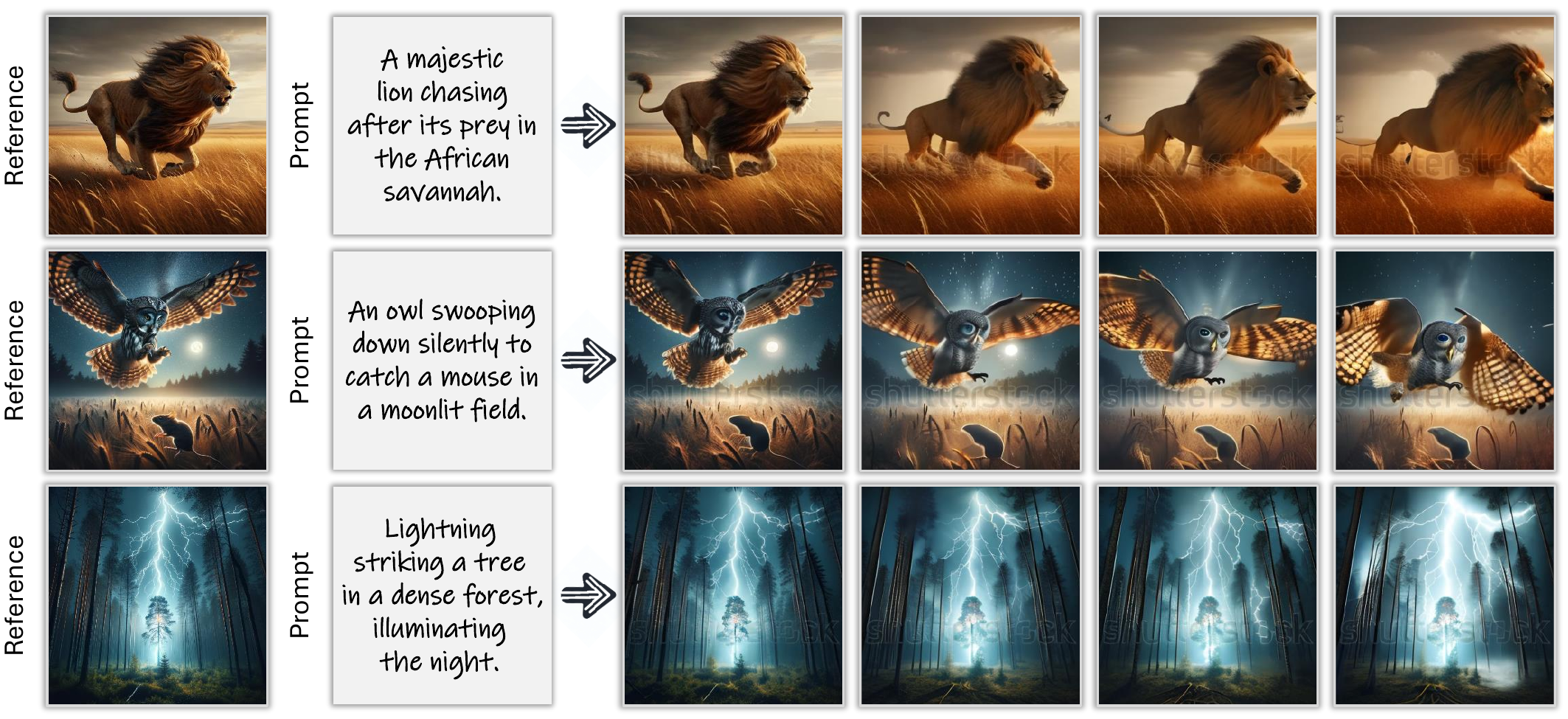}
    \vspace{-8mm}
    \captionsetup{type=figure}
    \caption{
    Exemplary results of text-image-to-video generation using our proposed approach, ART$\bigcdot$V. Our method skillfully captures object motion while preserving the overall scene, showcasing rich details and maintaining a high level of aesthetic quality. Reference images are generated by DALL-E 3 \cite{dalle2023}.
    }
    \label{fig:teaser-ti2v}
    \vspace{-2mm}
\end{center}
}]

\blfootnote{$^{\ast}$~This work is done when the author is an intern with MSRA.}
\blfootnote{$^{\dagger}$~Corresponding author.}

\begin{abstract}

We present ART$\bigcdot$V, an efficient framework for auto-regressive video generation with diffusion models. Unlike existing methods that generate entire videos in one-shot, ART$\bigcdot$V generates a single frame at a time, conditioned on the previous ones. The framework offers three distinct advantages. First, it only learns simple continual motions between adjacent frames, therefore avoiding modeling complex long-range motions that require huge training data. Second, it preserves the high-fidelity generation ability of the pre-trained image diffusion models by making only minimal network modifications. Third, it can generate arbitrarily long videos conditioned on a variety of prompts such as text, image or their combinations, making it highly versatile and flexible. To combat the common drifting issue in AR models, we propose masked diffusion model which implicitly learns which information can be drawn from reference images rather than network predictions, in order to reduce the risk of generating inconsistent appearances that cause drifting. Moreover, we further enhance generation coherence by conditioning it on the initial frame, which typically contains minimal noise. This is particularly useful for long video generation. When trained for only two weeks on four GPUs, ART$\bigcdot$V already can generate videos with natural motions, rich details and a high level of aesthetic quality. Besides, it enables various appealing applications, \eg composing a long video from multiple text prompts.

\end{abstract}    
\vspace{-4mm}
\section{Introduction}
\label{sec:intro}

Recently, text-to-image (T2I) generation~\cite{rombach2022high,midjourney2022,dalle2023} has been significantly advanced by generative diffusion models~\cite{sohl2015deep, ho2020denoising, nichol2021improved, song2019generative, song2020score} and large scale text-image datasets such as Laion5B~\cite{schuhmann2022laion}. The success has also catalyzed a remarkable proliferation of research in text-to-video (T2V) generation \cite{wu2023tune, blattmann2023align, ho2022imagen, singer2022make, luo2023videofusion, wang2023videocomposer, xing2023make, wang2023videofactory, zhou2022magicvideo, he2022latent, esser2023structure, an2023latent, chen2023control, zhang2023controlvideo, xing2023simda, fei2023empowering, ho2022video, gu2023reuse, wang2023modelscope, wang2023lavie, zhang2023show, zhao2023motiondirector, qiu2023freenoise, li2023videogen, ge2023preserve, chen2023videocrafter1, chen2023seine},  driven by the intrinsic allure of the potential breakthroughs.

Existing T2V methods~\cite{ho2022imagen,he2022latent} usually adopt a straightforward framework in which they generate entire videos at once using a spatial-temporal U-Net. However, they often produce videos with unrealistic motions. This is because learning the long-range motions is a highly ambiguous and complex task, which requires a significantly larger training dataset than that used in T2I, such as Laion5B~\cite{schuhmann2022laion}, which unfortunately is prohibitively expensive to collect and train on. Even the largest video dataset available~\cite{bain2021frozen} represents only a fraction of Laion5B. Therefore, we argue that achieving the ``stable diffusion'' moment in T2V using this framework is difficult.

\begin{figure}[t!]
    \centering
    \vspace{-2mm}
    \includegraphics[width=0.48\textwidth]{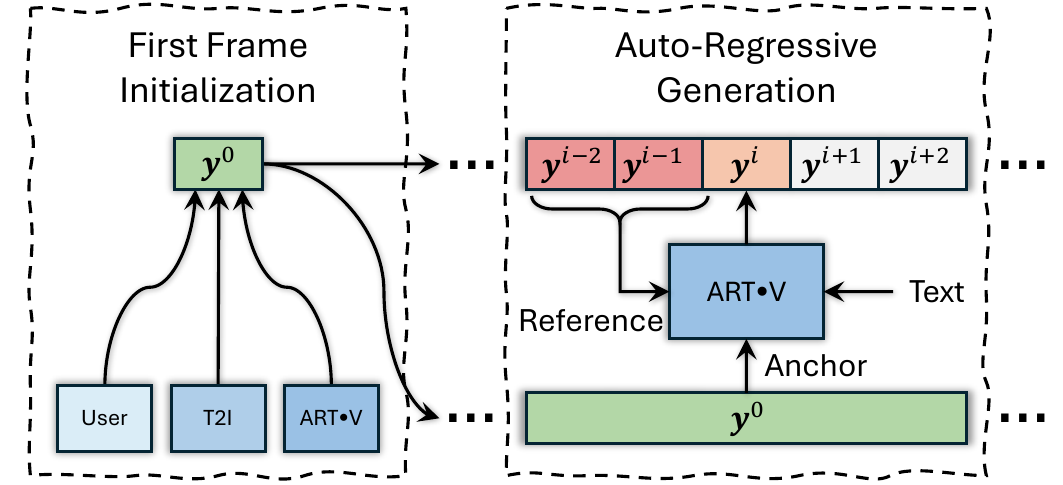}
    \vspace{-6mm}
    \caption{Overview of our video generation system ART$\bigcdot$V, consisting of first frame initialization process and auto-regressive generation process. The first frame can be initialized by users, T2I models \cite{rombach2022high,midjourney2022,dalle2023} or our ART$\bigcdot$V itself.
    }
    \vspace{-4mm}
    \label{fig:overview}
\end{figure}

In this work, we present ART$\bigcdot$V, a framework that generates video frames auto-regressively. As shown in \cref{fig:overview}, it first obtains a key frame as initialization. Then, with the key frame, or multiple copies of it, depending on the length of the conditioning sequence, ART$\bigcdot$V generates subsequent frames auto-regressively, one frame at a time. 
The conditioning frames, typically one or two previous frames, are concatenated and injected into a pre-trained image diffusion model~\cite{rombach2022high} using T2I-Adapter~\cite{mou2023t2i} (similar as ControlNet \cite{zhang2023adding} but smaller), for conditional generation. The resulting model is more efficient compared to previous methods, as it only needs to learn simple continuous motions between adjacent frames. Besides, it minimizes alternations to the pre-trained image diffusion model, eliminating the necessity for additional temporal layers, and preserving its high-fidelity generation capability. Contrary to conventional wisdom,  our auto-regressive model matches the inference speed of one-shot video models, while facilitating larger batch sizes during training.

\begin{figure*}[t!]
    \centering
    \vspace{-2mm}
    \includegraphics[width=0.9\textwidth]{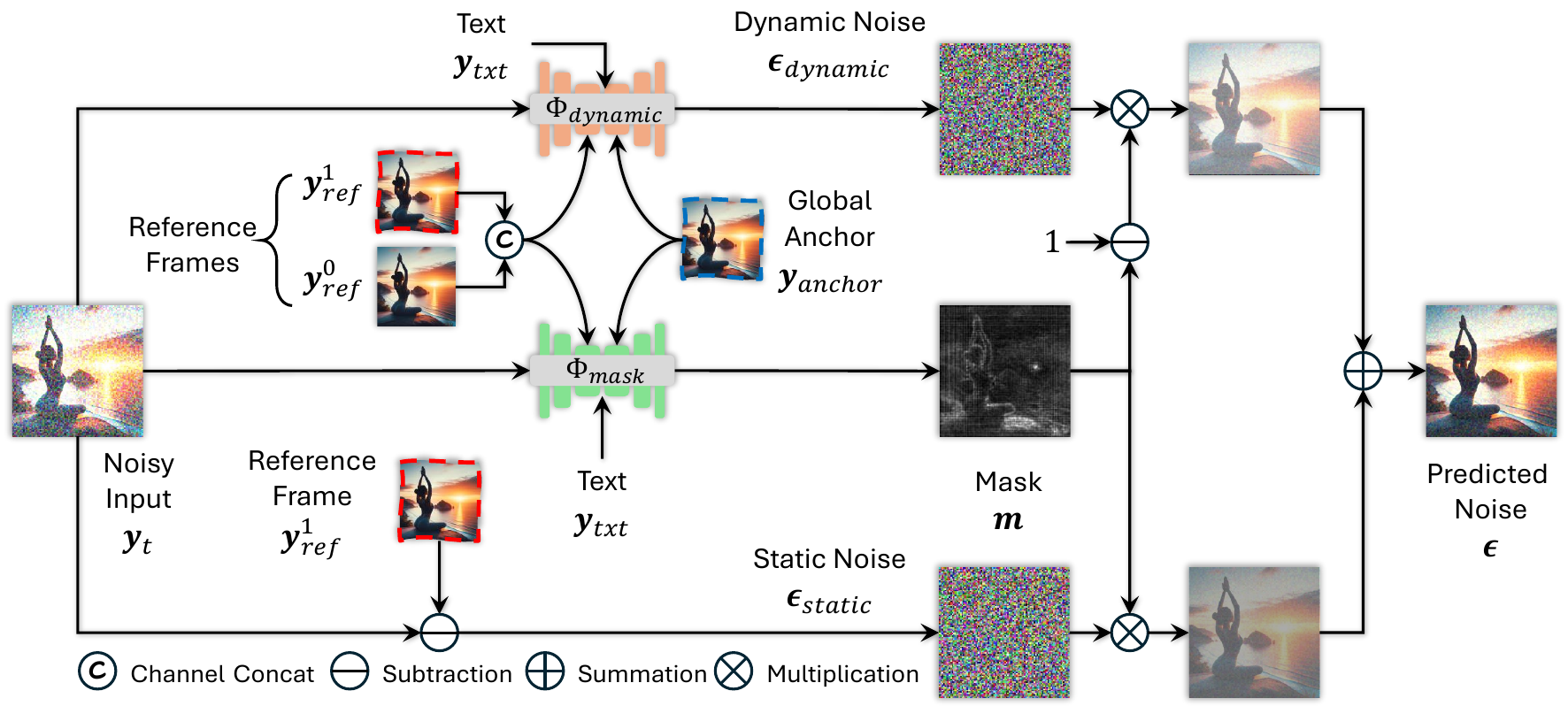}
    \vspace{-2mm}
    \caption{Illustration of the proposed masked diffusion model (MDM), conditioned on text, two reference frames and a global anchor frame. The predicted noise of MDM is composed of dynamic noise and static noise, which are scaled by a predicted mask. We employ two sub-networks $\Phi_{dynamic}$ and $\Phi_{mask}$ to predict dynamic noise and mask, respectively. Static noise is directly derived by subtraction of noisy input and reference frame. We initialize $\Phi_{dynamic}$ with Stable Diffusion 2.1 \cite{rombach2022high}, while $\Phi_{mask}$ is randomly initialized. Reference frames and global anchor frame are injected into two sub-networks by using T2I-Adapter \cite{mou2023t2i} and cross attention \cite{rombach2022high}, respectively. 
    Notably, the diffusion process is conducted in the latent space as in \cite{rombach2022high}. The autoencoder is omitted here for brevity.
    }
    \vspace{-3mm}
    \label{fig:mask_diffusion}
\end{figure*}

To combat drifting in AR models, we propose masked diffusion, which learns a mask that determines which information can be directly drawn from reference images, rather than from network predictions, to reduce the chance of generating inconsistent appearance. The static noise, obtained by subtracting the reference image from the input noised image, is a short-cut to propagate reference images to the diffusion model. Therefore, the network only needs to predict the remaining part of the noise, which we call as dynamic noise. \cref{fig:mask_diffusion} shows an overview of the proposed masked diffusion. Moreover, we further enhance the generation process by conditioning it on the initial frame, which sets the tone for the overall scene and appearance details, further promoting global coherence. We call the above scheme anchored conditioning, benefiting long video generation as well. Finally, we perform noise augmentation to the reference frames to bridge the gap between training and testing. We combine the above techniques to arrive at ART$\bigcdot$V, which effectively mitigates the drifting issue.

We train our model on five million text-video pairs filtered from the WebVid-10M dataset~\cite{bain2021frozen}. Due to limited GPU resouces, we only train the model for two weeks on four A100 GPUs. However, we find that ART$\bigcdot$V can already generate videos with natural motions, rich details and a high level of aesthetic quality. 
Though trained on low-resolution data, ART$\bigcdot$V can directly generate impressive high-resolution videos, as shown in Fig.~\ref{fig:exp-ti2v}.
It also achieves better quantitative results than the previous methods (they only represent proof-of-concept results since the methods are not fairly comparable due to differences in model size, training data and GPU resources). \cref{fig:teaser-ti2v} shows some examples. Most importantly, the simplicity of our model makes it highly scalable to larger training data and longer training time, which we believe can further improve the results. Besides, ART$\bigcdot$V enables various appealing applications. For example, it can generate long videos from multiple text prompts for story telling. It can also animate single images based on descriptive texts.

\section{Related Work}
\label{sec:related}

\noindent \textbf{Text-to-Video Generation.} The problem has seen remarkable progress recently.
Early T2V models demonstrated the possibility of generating videos in simple close-set domain \cite{mittal2017sync, pan2017create, marwah2017attentive, li2018video, gupta2018imagine, liu2019cross} and further exploited Transformer-based model \cite{vaswani2017attention} to achieve open-domain generation \cite{wu2021godiva, wu2022nuwa, hong2022cogvideo, villegas2022phenaki}.
Recently, diffusion-based T2V systems \cite{wu2023tune, blattmann2023align, ho2022imagen, singer2022make, luo2023videofusion, wang2023videocomposer, xing2023make, wang2023videofactory, zhou2022magicvideo, he2022latent, esser2023structure, an2023latent, chen2023control, zhang2023controlvideo, xing2023simda, fei2023empowering, ho2022video, gu2023reuse, wang2023modelscope, wang2023lavie, zhang2023show, zhao2023motiondirector, qiu2023freenoise, li2023videogen, ge2023preserve, chen2023videocrafter1, chen2023seine, ramesh2022hierarchical} have shown groundbreaking progress.
Models like ModelScope \cite{wang2023modelscope} and Imagen Video \cite{ho2022imagen} trained T2V models from scratch, demanding a huge text-video dataset and numerous GPU resources which is prohibitive for most cases.
In contrast, most works \cite{singer2022make, wu2023tune, blattmann2023align, luo2023videofusion, li2023videogen, zhang2023show, wang2023lavie, xing2023simda, wang2023videofactory} leveraged T2I model priors such as Stable Diffusion \cite{rombach2022high} for T2V by freezing or finetuning the pre-trained weights, showcasing compelling results.
However, these methods, usually generating entire videos in one-shot, suffer from generating unrealistic large motions or very limited motions.
In this work, we propose ART$\bigcdot$V, a generation system that avoids the challenge of learning complex long-range motion via auto-regressive first-order motion prediction, facilitating efficient training. 



\noindent \textbf{Auto-Regressive Video Generation.} This is a burgeoning research area that aims to generate realistic and coherent videos by predicting each frame based on previously generated frames.
Generally, three strategies have been employed. 
The first is pixel-level auto-regression.
Some representative methods attempt to estimate the joint distribution of pixel value auto-regressively \cite{kalchbrenner2017video}, speed up the processing by realizing a parallelized PixelCNN \cite{reed2017parallel}, and scale the techniques of auto-regressive Transformer architectures \cite{vaswani2017attention} to accommodate modern hardware accelerators \cite{weissenborn2019scaling}. 
The second is frame-level auto-regression.
Huang \etal \cite{huang2022single} proposed auto-regressive GAN to predict frames based on a single still frame.
By overcoming error accumulation problem of AR, the complementary masking is introduced to promote the generation quality.
The third is latent-level auto-regression, which significantly saves processing time due to reduced data redundancy and achieves a good time-quality trade-off \cite{walker2021predicting, rakhimov2020latent, yan2021videogpt, seo2022harp}.
Our ART$\bigcdot$V generation system, belonging to latent-level auto-regression, is the first attempt exploiting auto-regressive framework in the context of T2V with diffusion models.


\section{Method}
\label{sec:method}

\subsection{System Overview}
\cref{fig:overview} shows an overview of ART$\bigcdot$V.
Given a text prompt $\boldsymbol{y}_{txt}$ and an optional reference frame $\boldsymbol{y}^0$, it generates a video $\mathbf{V} = \{\boldsymbol{y}^0, \boldsymbol{y}^1, ...,\boldsymbol{y}^i,...\boldsymbol{y}^N\}$. If $\boldsymbol{y}^0$ is not available, the system can use existing T2I models to generate one, or uses ART$\bigcdot$V itself to generate one conditioned on blank images.  

It trains a conditional diffusion model $\Phi(\cdot;\theta)$ parameterized by $\theta$ to perform auto-regressive generation, which is formulated as
\begin{equation}
\label{equ:1}
\begin{aligned}
\boldsymbol{y}^i &= \Phi\left( \boldsymbol{y}_{txt}, \mathcal{R}^{i}; \theta\right), \\
\end{aligned}
\vspace{-2mm}
\end{equation}
where $\mathcal{R}^i$ denotes the set of conditional frames for generating $\boldsymbol{y}^i$. In implementation, $\mathcal{R}^i$ includes the previous two frames and an global anchor frame, denoted as $\boldsymbol{y}^{i-1}_{ref}$, $\boldsymbol{y}^{i-2}_{ref}$ and $\boldsymbol{y}_{anchor}$, to encode first-order motions. 

Our model is built on Stable Diffusion 2.1 \cite{rombach2022high} (SD2.1). To support image conditional generation, the two reference frames are concatenated along the channel dimension and injected into SD2.1 in a T2I-Adapter \cite{mou2023t2i} style, while the global anchor frame adopts cross attention for injection. We do not introduce additional temporal modeling modules such as 3D convolutions and attention layers, which are required by previous T2V models. This is because we only need to model short motions between adjacent frames. In the following, we will elaborate our proposed techniques for alleviating the drifting issue in AR models.

\subsection{Masked Diffusion Model (MDM)}

In standard diffusion process, all pixels are predicted from random noises by networks which have large chance of generating appearances inconsistent with the previous frames. As prediction proceeds auto-regressively, the accumulated errors will eventually lead to drifting. The core idea of MDM is to implicitly learn a mask determining which information can be drawn directly from closely related conditional images rather than network predictions to reduce inconsistency. \cref{fig:mask_diffusion} shows an overview of MDM.

\begin{figure}[!t]
    \centering
    \includegraphics[width=0.49\textwidth]{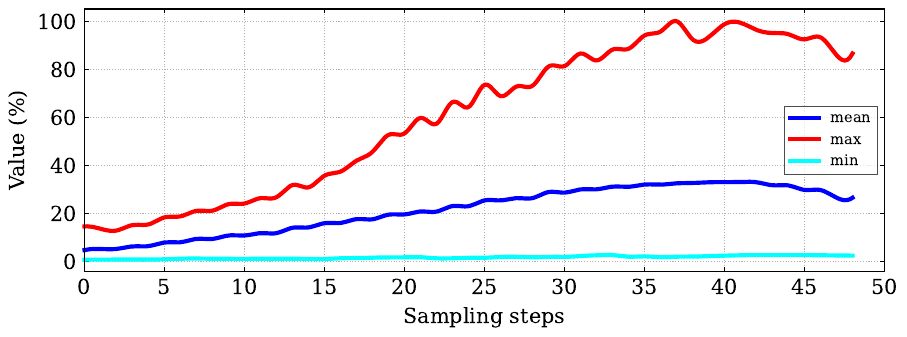}
    \vspace{-7mm}
    \caption{Value distribution of the estimated mask by mask diffusion model during different sampling steps. The maximum sampling step is 50.
    }
    \vspace{-1mm}
    \label{fig:ablation_mask}
\end{figure}

As shown in \cref{fig:mask_diffusion}, MDM has two U-Nets for predicting noise and mask, respectively. The static noise, directly obtained by subtracting the reference image from the input noised image, is a short-cut to propagate information in the reference image to the diffusion process. We find that the model tends to copy more from reference images at later denoising steps, which effectively reduces the risk of generating inconsistent high-frequency appearances that cause drifting. This is illustrated in \cref{fig:ablation_mask}.
The U-Net hence only needs to predict the remaining part of the noise, which we call as dynamic noise. In the following, we will formally introduce the method.

\paragraph{Diffusion Model Preliminaries.} 
Diffusion model has a forward and a backward process, respectively. The forward process gradually adds noises to the clean data $\boldsymbol{y}_0 \sim q\left(\boldsymbol{y}_0\right)$, which can be formulated as:
\begin{equation}
\label{equ:2}
\begin{aligned}
q\left(\boldsymbol{y}_t \mid \boldsymbol{y}_{t-1}\right)=\mathcal{N}\left(\boldsymbol{y}_t ; \sqrt{1-\beta_t} \boldsymbol{y}_{t-1}, \beta_t \mathbf{I}\right),
\end{aligned}
\end{equation}
where $t \in \{1,...,T\}$ and $\beta_t \in (0, 1)$ is a fixed variance schedule.
Denote that $\alpha_t=1-\beta_t$ and $\bar{\alpha}_t=\prod_{i=1}^t \alpha_i$, we can directly sample $\boldsymbol{y}_t$ in a closed form from the distribution $q(\boldsymbol{y}_t | \boldsymbol{y}_0)$ at an arbitrary timestep $t$: 
\begin{equation}
\label{equ:3}
\begin{aligned}
\boldsymbol{y}_t = \sqrt{\Bar{\alpha_t}}\boldsymbol{y}_0 + \sqrt{1-\bar{\alpha_t}}\boldsymbol{\epsilon},
\end{aligned}
\end{equation}
where $\boldsymbol{\epsilon} \sim \mathcal{N}(\mathbf{0}, \mathbf{I})$.

The backward process reverses the forward process, which eventually maps Gaussian noises $\boldsymbol{y}_T \sim \mathcal{N}(\mathbf{0}, \mathbf{I})$ to the target data. Specifically, the backward denoising process solves the posterior $q(\boldsymbol{y}_{t-1}| \boldsymbol{y}_t)$, which can be approximated by training a deep neural network $\Phi(\cdot;\theta)$ to predict the noise $\boldsymbol{\epsilon}$ added to the data.
The training objective is formulated as:
\begin{equation}
\label{equ:4}
\begin{aligned}
\mathbb{E}_{\boldsymbol{y}, \boldsymbol{\epsilon} \sim \mathcal{N}(\mathbf{0}, \mathbf{I}), t}\left[\left\|\boldsymbol{\epsilon}-\Phi\left(\boldsymbol{y}_t, \boldsymbol{c}, t ; \theta\right)\right\|_2^2\right],
\end{aligned}
\end{equation}
where $\mathbf{c}$ denotes the conditions that represent the reference and global anchor frames, and texts in our ART$\bigcdot$V system.

\paragraph{Mask Prediction and Dynamic Noise.}

In MDM, noise prediction in \cref{equ:4} is realized by two networks: dynamic noise prediction network $\Phi_{dynamic}(\cdot;\theta_0)$ and mask prediction network $\Phi_{mask}(\cdot;\theta_1)$.
Without loss of generality, we define $\sigma=\sqrt{\bar{\alpha_t}}$ and $\lambda=\sqrt{1-\bar{\alpha_t}}$.
We omit $t$ for brevity.
We reformulate
\cref{equ:3} as:
\begin{equation}
\label{equ:5}
\begin{aligned}
\boldsymbol{y}_t &= \sigma \boldsymbol{y}_0 + \lambda\boldsymbol{\epsilon} \\
&= (\boldsymbol{y}_{ref} + \boldsymbol{y}_{res}) + \lambda\boldsymbol{\epsilon} \\
&= \boldsymbol{y}_{ref} + (\boldsymbol{y}_{res} + \lambda\boldsymbol{\epsilon}) \\
&= \boldsymbol{y}_{ref} + \boldsymbol{\epsilon}^{\prime},
\end{aligned}
\end{equation}
where $\boldsymbol{y}_{ref}$ is the reference frame, and $\boldsymbol{y}_{res}$ denotes the residual component between $\sigma \boldsymbol{y}_0$ and $\boldsymbol{y}_{ref}$.
Therefore, the $\boldsymbol{\epsilon}$, which needs to be predicted by the diffusion model $\Phi(\cdot;\theta)$  in \cref{equ:4}, can be derived from \cref{equ:5}:
\begin{equation}
\label{equ:6}
\begin{aligned}
\boldsymbol{\epsilon} &=  \frac{\boldsymbol{y}_{ref}+\boldsymbol{\epsilon}^{\prime}-\sigma\boldsymbol{y}_0}{\lambda} \\
&= \frac{\boldsymbol{y}_{ref}-\sigma\boldsymbol{y}_0}{\lambda} + \frac{\boldsymbol{\epsilon}^{\prime}}{\lambda} \\
&= \frac{\boldsymbol{\epsilon}^{\prime \prime}}{\lambda} + \frac{\boldsymbol{\epsilon}^{\prime}}{\lambda} \\
&= \boldsymbol{\epsilon}_{static} + \boldsymbol{\epsilon}_{dynamic},
\end{aligned}
\end{equation}
where $\boldsymbol{\epsilon}_{static}$ and $\boldsymbol{\epsilon}_{dynamic}$ represents the static noise and dynamic noise, respectively.

We can see from \cref{equ:5} and \cref{equ:6} that the static noise $\boldsymbol{\epsilon}_{static}$ is from the reference image $\boldsymbol{y}_{ref}$, which can be directly propagated to the output and is expected to mitigate error accumulation. 
In our implementation, we make approximation $\boldsymbol{\epsilon}_{static} \simeq \boldsymbol{y}_{ref} - \boldsymbol{y}_t$.
In such a way, $\boldsymbol{\epsilon}_{static}$ can be directly derived from reference images and noised input input, which do not need to be predicted.
The dynamic noise $\boldsymbol{\epsilon}_{dynamic}$ contains the residual component $\boldsymbol{y}_{res}$ that changes dynamically, which needs to be predicted by our noise prediction network $\Phi_{dynamic}(\cdot;\theta_0)$.
In order to determine the contributions of static and dynamic noises, we employ the mask prediction network $\Phi_{mask}(\cdot;\theta_1)$ to predict a mask $\boldsymbol{m}$.
Eventually, the final predicted noise of our mask diffusion model is obtained by:
\begin{equation}
\label{equ:7}
\begin{aligned}
\widehat{\boldsymbol{\epsilon}} = \boldsymbol{m}\cdot\boldsymbol{\epsilon}_{static} + (1-\boldsymbol{m})\cdot\boldsymbol{\epsilon}_{dynamic}.
\end{aligned}
\end{equation}
The two networks can be optimized by \cref{equ:4}.

\begin{table*}[t!]
    \setlength{\tabcolsep}{8pt}
    \renewcommand{\arraystretch}{1}
    \begin{center}
    \caption{Quantitative comparisons with SoTA for zero-shot video generation on UCF-101 \cite{soomro2012ucf101} and MSR-VTT \cite{xu2016msr}.}
    \vspace{-2mm}
    \label{tab:main}
    \resizebox{1.0\linewidth}{!}{
    \begin{tabular}{lccccccccc}
    \toprule[0.8pt]
    \multirow{2}{*}{Methods} & \multirow{2}{*}{Training Data}  &\vline& \multicolumn{3}{c}{UCF-101\cite{soomro2012ucf101}} &\vline & \multicolumn{3}{c}{MSR-VTT\cite{xu2016msr}} \\
    \cline{4-6}  \cline{8-10}
    & & \vline& Zero-shot & FVD $\downarrow$ & IS $\uparrow$ & \vline& Zero-shot & FVD $\downarrow$ & CLIPSIM $\uparrow$ \\
    \cline{1-2} \cline{4-6}  \cline{8-10}
    GODIVA \cite{wu2021godiva} & MSR-VTT \cite{xu2016msr} & \vline& Yes & - & - & \vline& No & - & 0.2402 \\
    NUWA \cite{wu2022nuwa} & MSR-VTT \cite{xu2016msr} & \vline& Yes & - & - & \vline& No & - & 0.2439 \\
    Make-A-Video \cite{singer2022make} & WebVid-10M \cite{bain2021frozen} + HD-VILA-100M \cite{xue2022advancing} & \vline& Yes & 367.23 & 33.00 & \vline& Yes & - & 0.3049 \\
    VideoFactory \cite{wang2023videofactory} & WebVid-10M \cite{bain2021frozen} + HD-VG-130M \cite{wang2023videofactory} & \vline& Yes & 410.00 & - & \vline& Yes & - & 0.3005 \\
    ModelScope \cite{wang2023modelscope} & WebVid-10M \cite{bain2021frozen} + LAION-5B \cite{schuhmann2021laion} & \vline& Yes & 410.00 & - & \vline& Yes & 550.00 & 0.2930 \\
    VideoGen \cite{li2023videogen} & WebVid-10M \cite{bain2021frozen} + Private-HQ-2K \cite{li2023videogen} & \vline& Yes & 554.00 & 71.61 & \vline& Yes & - & 0.3127 \\
    Lavie \cite{wang2023lavie} & WebVid-10M \cite{bain2021frozen} + LAION-5B \cite{schuhmann2021laion} & \vline& Yes & 526.30 & - & \vline& Yes & - & 0.2949 \\
    VidRD \cite{gu2023reuse} & WebVid-2M \cite{bain2021frozen} + TGIF \cite{li2016tgif} + VATEX \cite{wang2019vatex} + Pexels \cite{pevels} & \vline& Yes & 363.19 & 39.37 & \vline& Yes & - & - \\
    PYoCo \cite{ge2023preserve} & Private-data \cite{ge2023preserve} & \vline& Yes & 355.19 & 47.76 & \vline& Yes & - & 0.3204 \\
    \cline{1-2} \cline{4-6}  \cline{8-10}
    LVDM \cite{he2022latent} & WebVid-2M \cite{bain2021frozen} & \vline& Yes & 641.80 & - & \vline& Yes & 742.00 & 0.2381 \\
    CogVideo \cite{hong2022cogvideo} & WebVid-5.4M \cite{bain2021frozen} & \vline& Yes & 702.00 & 25.27 & \vline& Yes & 1294.00 & 0.2631 \\
    MagicVideo \cite{zhou2022magicvideo} & WebVid-10M \cite{bain2021frozen} & \vline& Yes & 699.00 & - & \vline& Yes & 998.00 & - \\
    Video-ldm \cite{blattmann2023align} & WebVid-10M \cite{bain2021frozen} & \vline& Yes & 550.61 & 33.45 & \vline& Yes & - & 0.2929 \\
    VideoComposer \cite{wang2023videocomposer} & WebVid-10M \cite{bain2021frozen} & \vline& Yes & - & - & \vline& Yes & 580.00 & 0.2932 \\
    VideoFusion \cite{luo2023videofusion} & WebVid-10M \cite{bain2021frozen} & \vline& Yes & 639.90 & 17.49 & \vline& Yes & 581.00 & 0.2795 \\
    SimDA \cite{xing2023simda} & WebVid-10M \cite{bain2021frozen} & \vline& Yes & - & - & \vline& Yes & 456.00 & 0.2945 \\
    \cline{1-2} \cline{4-6}  \cline{8-10}
    ART$\bigcdot$V + W/O Image (Ours) & WebVid-5M \cite{bain2021frozen} & \vline& Yes & 567.20 & 26.89 & \vline& Yes & 356.50 & 0.2897 \\
    ART$\bigcdot$V + SDXL \cite{podell2023sdxl} (Ours) & WebVid-5M \cite{bain2021frozen} & \vline& Yes & 539.57 & 36.21 & \vline& Yes & 413.01 & \textbf{0.3022} \\
    ART$\bigcdot$V + GT Image (Ours) & WebVid-5M \cite{bain2021frozen} & \vline& Yes & \textbf{315.69} & \textbf{50.34} & \vline& Yes & \textbf{291.08} & 0.2859 \\

    \bottomrule[0.8pt]
    \end{tabular}
    }
    \end{center}
    \vspace{-5mm}
\end{table*}

\begin{figure*}[!t]
    \centering
    \includegraphics[width=0.97\textwidth]{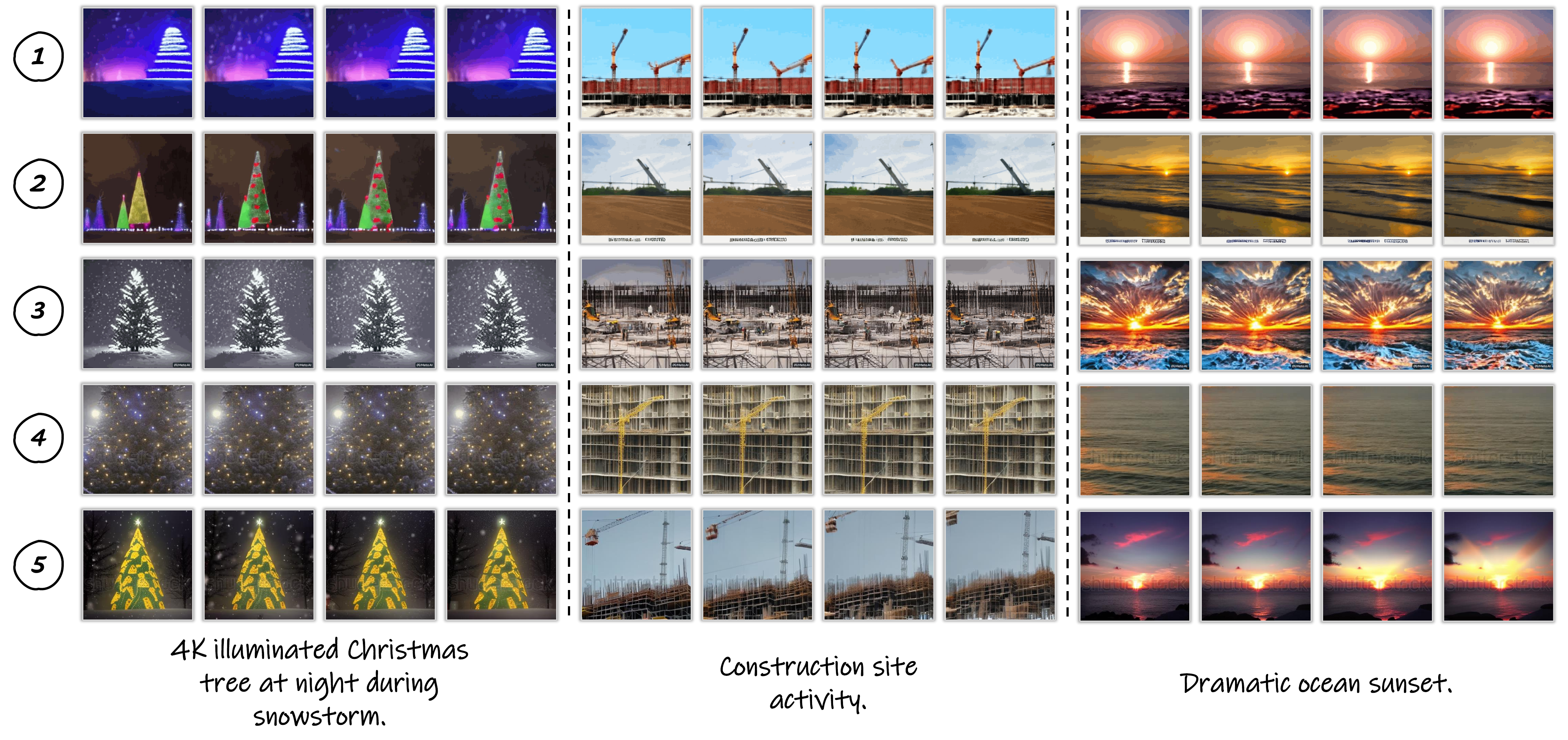}
    \vspace{-3.7mm}
    \caption{Visual comparisons of text-to-video generation. The results of row 1 to row 5 are sampled from VDM \cite{ho2022video}, CogVideo \cite{hong2022cogvideo}, Make-A-Video \cite{singer2022make}, ModelScope \cite{wang2023modelscope} and Our ART$\bigcdot$V.
    }
    \vspace{-3mm}
    \label{fig:exp-t2v}
\end{figure*}

\begin{figure*}[!t]
    \centering
    \includegraphics[width=0.95\textwidth]{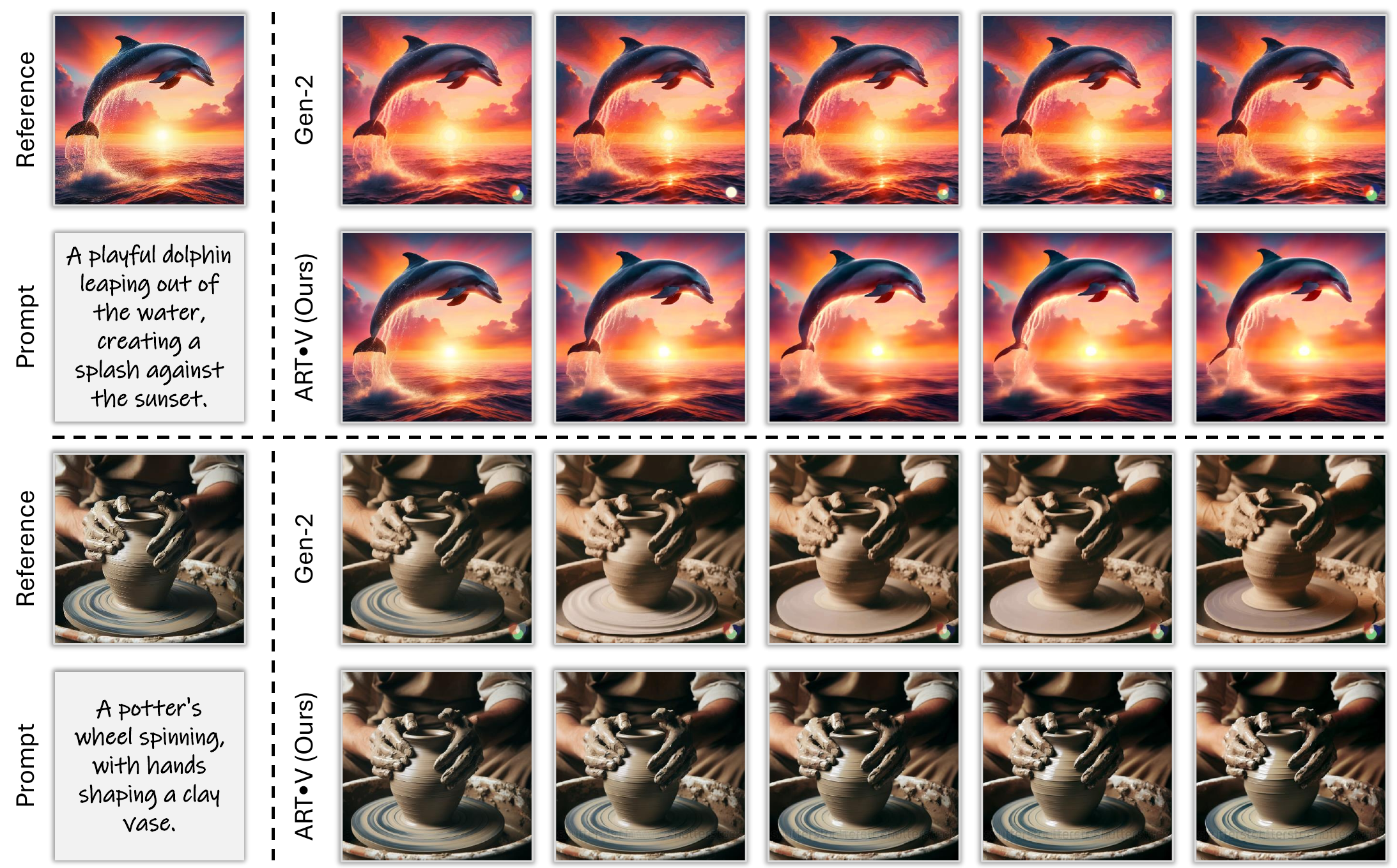}
    \vspace{-2mm}
    \caption{Visual comparisons of text-image-to-video generation. Reference image generated by DALL-E 3 \cite{dalle2023}.
    Notebly, ART$\bigcdot$V is trained on $320\times 320$ video data, while the inference is performed on $768\times 768$ in these cases.
    }
    \vspace{-2mm}
    \label{fig:exp-ti2v}
\end{figure*}

\subsection{Noise Augmentation}
Drifting issue in our ART$\bigcdot$V generation system arises not only from prediction error but also from train-test discrepancy. 
During training, the model utilizes ground truth frames as references and the global anchor. 
However, during testing, it conditions on generated frames prone to noises. Inspired by~\cite{rombach2022high}, we slightly corrupt reference and global anchor frames using the forward diffusion process in \cref{equ:3}. In particular, for each training step, we randomly sample a noise level $t \in [0, T_{max}]$. In such a way, the model has the chance to see clean reference frames and corrupted ones, respecting the case of inference and expected to address the error accumulation problem during inference.
Following~\cite{rombach2022high}, we also use the noise level $t_{\text{test}}$ as an additional condition by adding it to the time step embedding of diffusion model. In inference, we use a fixed noise level of $t_{\text{test}}=200$, validated by the ablation study in \cref{sec:analysis}.


\subsection{Anchored Conditioning}
In addition to using masked diffusion model and noise augmentation to address drifting issue in our ART$\bigcdot$V generation system, we introduce a novel design, anchored conditioning, expected to promote model capacity for long video generation. 
One key challenge in generating long videos is to maintain consistency in terms of scenes and objects throughout videos, solved by a global anchor in ART$\bigcdot$V.

In detail, we use the first frame, which is free from noises, as a stable anchor frame $\boldsymbol{y}_{anchor}$ to preserve the content, in whole videos.
In training, we randomly select one frame within a fixed time window range preceding the current one to serve as the global anchor frame. 
We expirically choose time window range as 10, to create relatively large motion variations.
We use cross attention \cite{vaswani2017attention} to inject the global anchor frame to the diffusion model.
The strategy addresses the inherent challenges in long text-to-video generation, providing a robust mechanism for faithfully retaining the scenes and objects.

\section{Experiment}
\label{sec:experiment}


\noindent \textbf{Datasets and Evaluation Metrics.}
To make quantitative and qualitative comparisons, we choose the publicly available datasets: WebVid-10M \cite{bain2021frozen}, MSR-VTT \cite{xu2016msr} and UCF-101 \cite{soomro2012ucf101}.
We split WebVid-10M to training subset and testing subset.
We make data cleaning on the training subset.
In specific, we use the public code \cite{motiondetect} to compute the motion score of each video and then only retain the videos whose motion scores are between $[1, 20]$.
Subsequently, we compute a CLIP score for each video and retain the top 5 million data that have largest CLIP scores \cite{radford2021learning}.
We train our model on this cleaned 5M dataset.
MSR-VTT \cite{xu2016msr} and UCF-101 \cite{soomro2012ucf101} are utilized for evaluation.
We report the Frechet Video Distance (FVD)  \cite{unterthiner2018towards}, Frechet Inception Distance (FID) \cite{parmar2022aliased}, Inception Score (IS) \cite{salimans2016improved} and CLIPSIM (average CLIP similarity between video frames and text) \cite{radford2021learning} for quantitative comparison.

\noindent \textbf{Implementation Details.}
We implement our method using Pytorch \cite{paszke2017automatic} and use AdamW solver for optimization.
We train our diffusion model with $1000$ noising steps and a linear noise schedule.
The exponential moving average (EMA) of model weights with $0.9999$ decay is adopted during training.
We set the learning rate as $1e^{-5}$ and keep it constant during the training process.
We use a batch size of 640.
For noise augmentation, we set the maximum noise level $T_{max}$ as $550$.
For inference, we employ classifier-free guidance \cite{ho2022classifier} to amplify the effect of the conditional signals of reference frames $\boldsymbol{y}_{ref}$, global anchor frame $\boldsymbol{y}_{anchor}$ and text prompts $\boldsymbol{y}_{text}$.
The guidance scales of $\boldsymbol{y}_{ref}$, $\boldsymbol{y}_{anchor}$ and $\boldsymbol{y}_{text}$ are set as $0.25$, $0.25$ and $6.5$, respectively.
During training, we randomly drop these conditions with a drop rate of $10\%$.

\subsection{Application}
We now demonstrate a wide range of applications of our ART$\bigcdot$V system.
Our ART$\bigcdot$V, only trained once without task-specific finetuning, can skillfully support multiple generation tasks.
In contrast, existing models like VideoCrafter1 \cite{chen2023videocrafter1} needs to train two individual models for T2V and TI2V, causing large training cost.

\noindent \textbf{Text-to-Video Generation.}
We first exploit our ART$\bigcdot$V to perform text-to-video generation, without the image provided by T2I models \cite{rombach2022high,midjourney2022,dalle2023} or users.
It is worth noting that, our model is trained using joint conditions of text and images.
Notably, we randomly drop the image condition with a drop rate of $10\%$ during training.
It suggests that the training cases of text-to-video take a small proportion.
However, we observe that, ART$\bigcdot$V, trained for text-image-to-video generation, is able to skillfully generate video by using text condition only.
Specifically, when there is no provided reference frames, we directly use our model to generate one from the text prompt, leaving the conditioned reference frames blank.
Then, we generate the subsequent frames conditioned on the generated reference frames. We demonstrate the quantitative results in \cref{tab:main}. We compare our method with the existing state-of-the-art methods on UCF-101 \cite{soomro2012ucf101} and MSR-VTT \cite{xu2016msr} in a zero-shot setting.
It can be clearly observed that, our method ART$\bigcdot$V, achieving FVD score of 567.20 and IS score of 26.89 in UCF-101, consistently outperforms existing methods such as VideoFusion \cite{luo2023videofusion}, MagicVideo \cite{zhou2022magicvideo}, LVDM \cite{he2022latent} and CogVideo \cite{hong2022cogvideo}.
In MSR-VTT, we keep the top performance in terms of FVD, and even outperform ModelScope \cite{wang2023modelscope} that utilizes additional high-quality datasets for training.
In \cref{fig:exp-t2v}, we also demonstrate some exemplary results of different methods using the same text prompts. The visual results also support the conclusions above, demonstrating the visually-satisfying results compared to the existing methods.
In addition, we believe if ART$\bigcdot$V is finetuned for T2V task, we will achieve better results.

\begin{figure}[!t]
    \centering
    \vspace{-3mm}
    \includegraphics[width=0.5\textwidth]{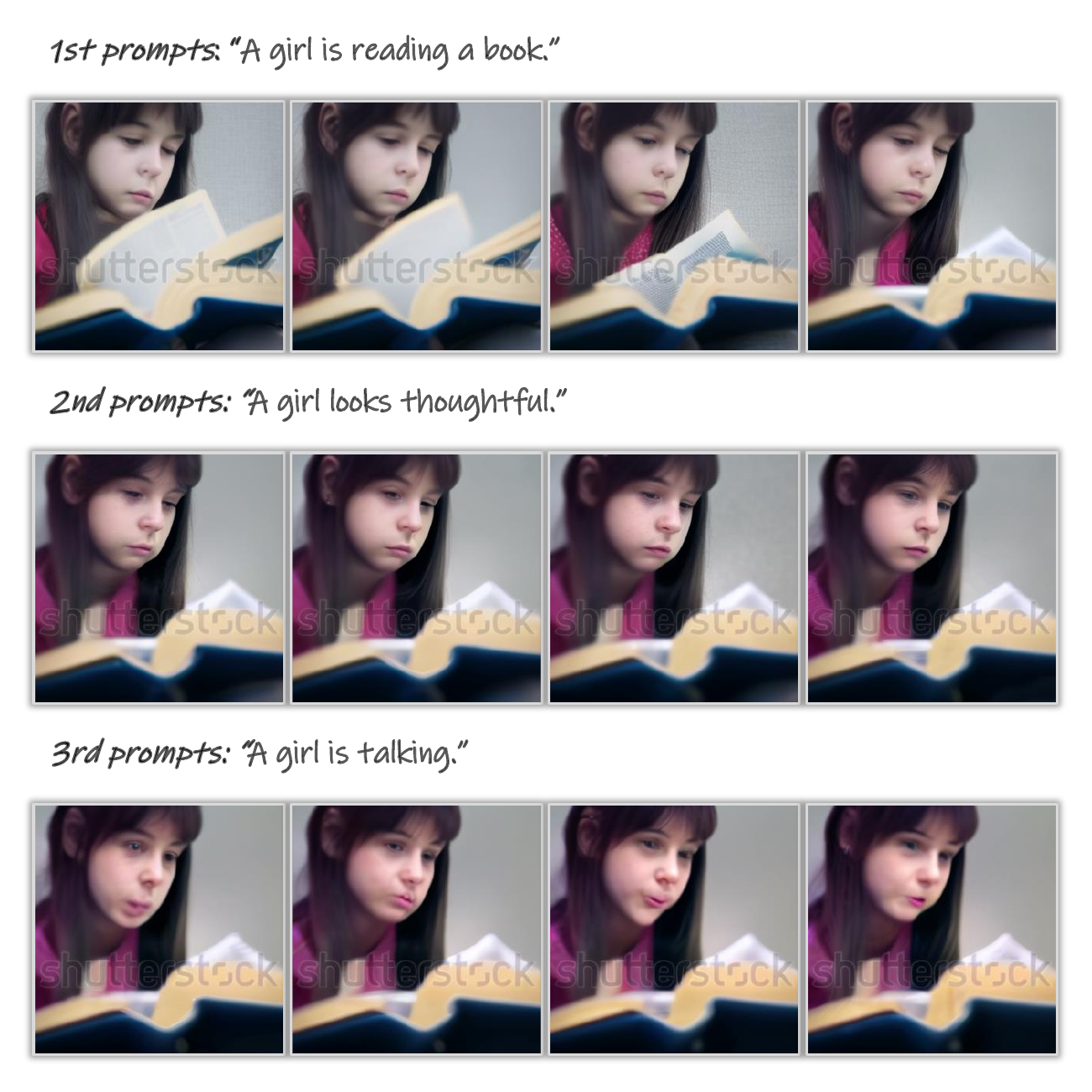}
    \vspace{-7mm}
    \caption{Visual result of multi-prompt long text-to-video generation. 16 frames are generated for each prompt.
    }
    \vspace{-5mm}
    \label{fig:exp-long-t2v}
\end{figure}

\noindent \textbf{Text-Image Conditioned Video Generation.}
ART$\bigcdot$V also offers the ability to animate a still image based on text prompts. We either employ the existing T2I models such as Stable Diffusion \cite{rombach2022high}, Midjourney \cite{midjourney2022}, and DALL-E 3 \cite{dalle2023} to generate reference images or directly use the images provided by users. 
The numbers are reported in \cref{tab:main}.
We make two variants of our method, termed ``ART$\bigcdot$V+SDXL" and ``ART$\bigcdot$V+GT Image", which utilize the image generated by SDXL \cite{podell2023sdxl} and GT image as the first frame, respectively.
As can be clearly observed, when conditioned on an additional image, our ART$\bigcdot$V achieves better results in terms of FVD and IS in UCF-101.
Especially, we achieve the SoTA results, FVD of 315.69 and IS of 50.34 in UCF-101, FVD of 291.08 in MSR-VTT, when GT image is taken as reference image. It demonstrates the superior performance of text-image conditioned generation of our ART$\bigcdot$V.

We demonstrate some visual exemplar videos generated by our method in \cref{fig:exp-ti2v}.
In these cases, ART$\bigcdot$V is exploited to generate high-resolution videos of $768\times768$, though the model is trained on $320\times320$.
We compare to a well-known video generation system Gen-2 \cite{gen2022} provided by a commercial company.
We generate the reference frame using DALLE 3 \cite{dalle2023}, which is then fed to ART$\bigcdot$V and Gen-2 to generate videos, respectively.

We observe that both ART$\bigcdot$V and Gen-2 are able to animate the given image using the text description, demonstrating good visual fidelity.
Notably, our ART$\bigcdot$V exhibits a superior ability to preserve appearance compared to Gen-2.
As can be seen from the second case of \cref{fig:exp-ti2v}, Gen-2 shows the severe color shifting problem, while our ART$\bigcdot$V preserves the content in the reference images, thanks to the proposed masked diffusion model and anchored conditioning.
In addition, the exceptional visual quality of ART$\bigcdot$V demonstrates that our method can achieve tuning-free, high-resolution video generation, thereby significantly reducing the training costs.
Nevertheless, Gen-2 show superior results in terms of visual detail and temporal consistency due to additional high-quality training data and temporal interpolation models, which is beyond the scope of this paper.
In contrast, we train ART$\bigcdot$V only using WebVid-5M, which has low resolution and quality.

\begin{figure}[t!]
    \centering
    \subcaptionbox{}{
    \includegraphics[width=0.255\textwidth]{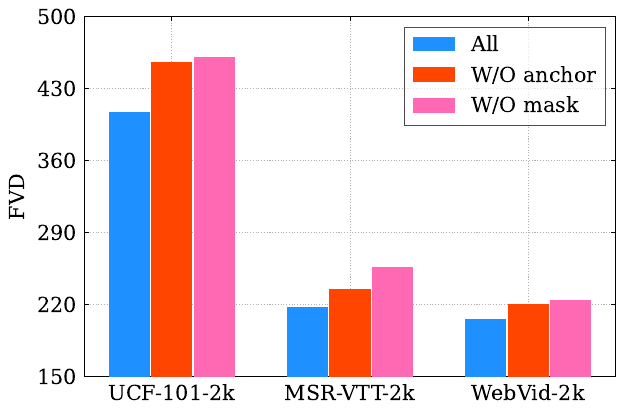}
    }
    \subcaptionbox{}{
    \includegraphics[width=0.196\textwidth]{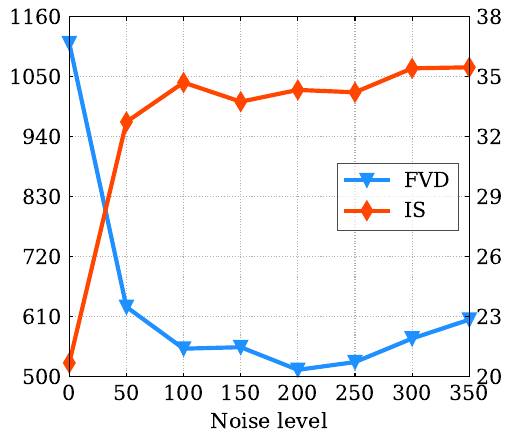}
    }
    \vspace{-3mm}
    \caption{(a) Ablation results of mask diffusion model and anchor conditioning on UCF-101-2k \cite{soomro2012ucf101}, MSR-VTT-2k \cite{xu2016msr} and WebVid-2k \cite{bain2021frozen}. 
    (b) Investigation results of noise augmentation on UCF-101-2k \cite{soomro2012ucf101}.
    }
\vspace{-5mm}
\label{fig:ablation_num}
\end{figure}

\noindent \textbf{Multi-Prompt Long Video Generation.}
ART$\bigcdot$V is suitable for long video generation due to its auto-regressive nature.
We can repeat the auto-regressive process to generate an arbitrarily long video, and require different segments of the video to be conditioned on different prompts. The leading frame of a video segment should be conditioned on the ending frames of last video segment to promote coherence and continuity. \cref{fig:exp-long-t2v} shows an example. We can see that our system can generate videos with coherent scenes and objects, and meanwhile the motions in each segment are faithful to the corresponding prompts.

\begin{figure}[!t]
    \centering
    \includegraphics[width=0.49\textwidth]{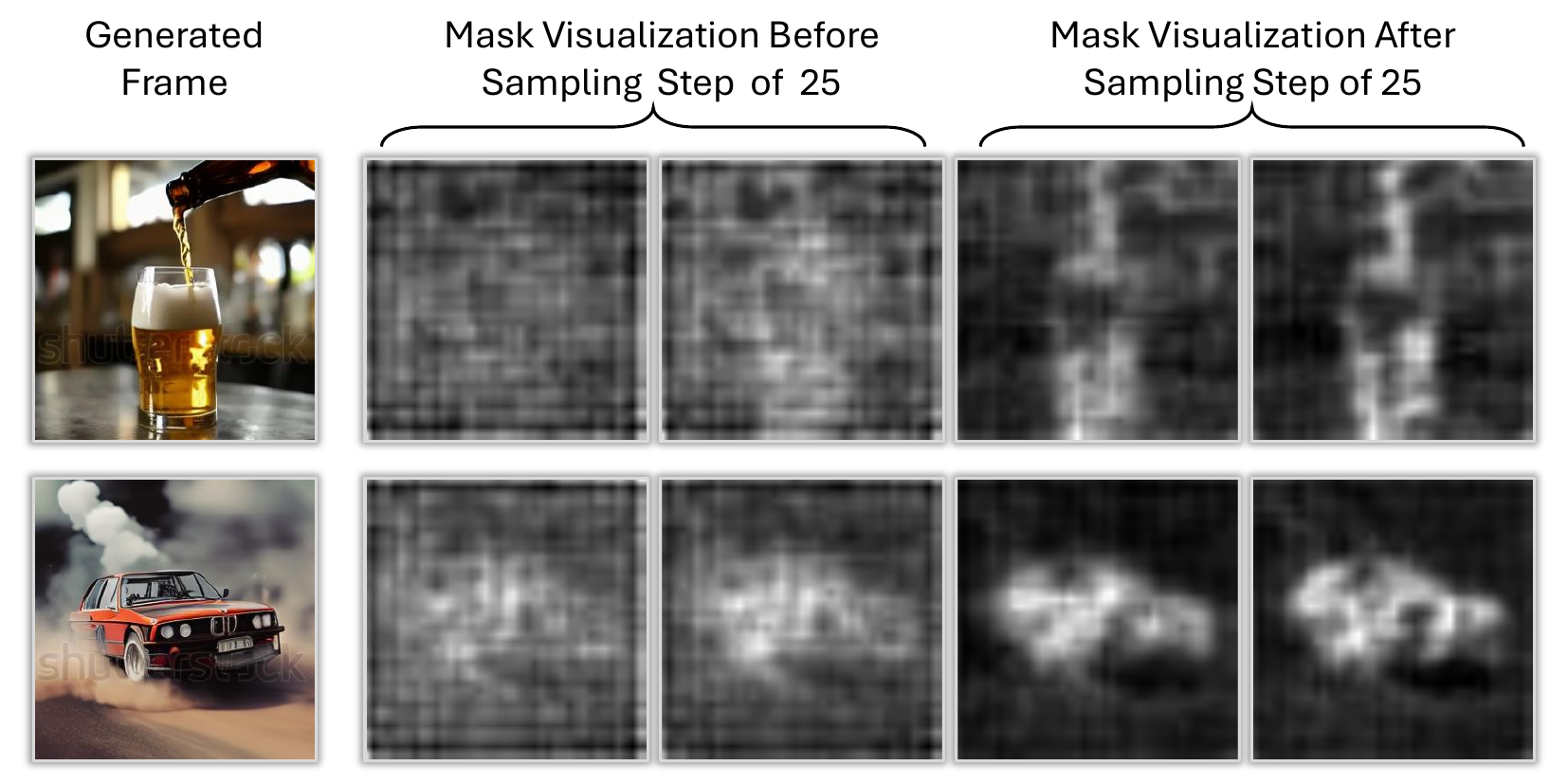}
    \vspace{-7mm}
    \caption{Visualization of estimated mask by mask diffusion model. Reference image is generated by SDXL \cite{podell2023sdxl}.
    }
    \vspace{-5mm}
    \label{fig:ablation_mask_vis}
\end{figure}

\subsection{Ablation Study}
\label{sec:analysis}

\begin{figure}[!t]
    \centering
    \includegraphics[width=0.5\textwidth]{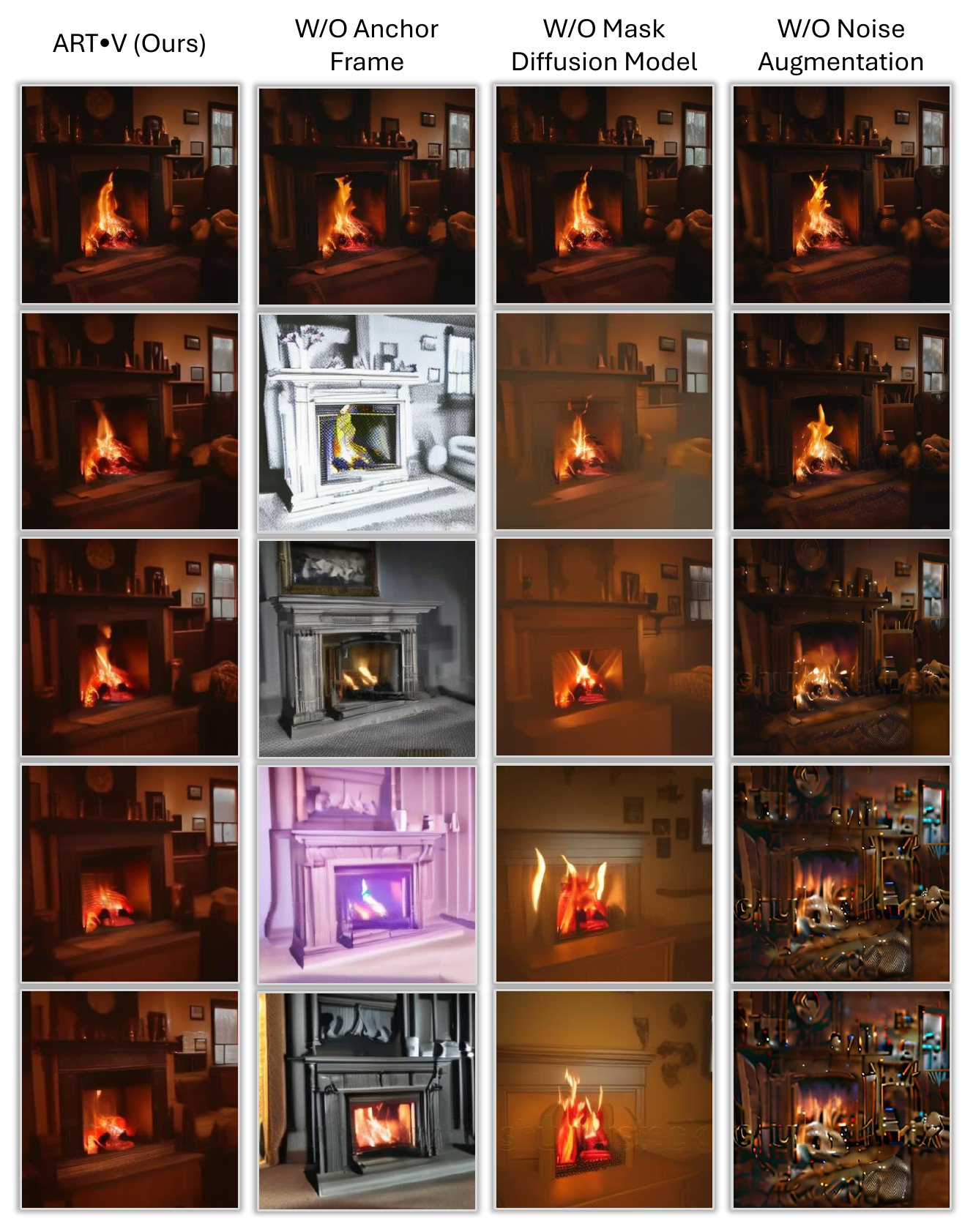}
    \vspace{-7mm}
    \caption{Visual results of ablation study. Reference image is generated by SDXL \cite{podell2023sdxl} by using prompt ``\textit{interior, fireplace.}''
    }
    \vspace{-5mm}
    \label{fig:ablation_vis}
\end{figure}

\noindent \textbf{Masked Diffusion Model.}
We propose masked diffusion model to alleviate the error accumulation in our ART$\bigcdot$V system.
To validate its effectiveness, we introduce a baseline which drops the mask prediction network.
It trains a single network to predict the noise in \cref{equ:7} as in standard diffusion models.
As shown in \cref{fig:ablation_num} (a), when we drop the masked diffusion, the performance drops significantly on all evaluation datasets.
We also visually compare the videos generated by different methods in ~\cref{fig:ablation_vis}. We can see that the model suffers from severe drifting without masked diffusion. In addition, the image quality is also degraded, losing many sharp details. These results demonstrate the importance of masked diffusion. 


We show the normalized strengths of the predicted masks at different time steps in \cref{fig:ablation_mask}. The average strength increases as the denoising step, suggesting that the diffusion model will use copy more from the reference images in later denoising steps, where diffusion models focus on high-frequency appearance generation~\cite{balaji2022ediffi}. So, our model will generate images that have similar appearance as the reference images, thus can effectively alleviate the drifting issue. 
\cref{fig:ablation_mask_vis} shows the masks at different denoising steps, which validates our conjecture.

\noindent \textbf{Noise Augmentation.}
In addition to masked diffusion model, we propose noise augmentation to further reduce error accumulation.
We investigate the effect of applying different levels of noises, \ie $t_{\text{test}}$, during inference. The numeric results are in \cref{fig:ablation_num} (b).
As can be observed, adopting noise augmentation brings significant performance boosts in terms of FVD and IS metrics.
When we increase the noise level, IS achieves consistently better results but the gains become marginal after exceeding 100.
In contrast, the FVD gets the best result when the noise level is 200 and shows performance drop when noise level exceeds 200. It is worth noting the value of 200 is approximately the average of the noise levels we applied during training.
\cref{fig:ablation_vis} shows the visual results of ablating noise augmentation, which is adversely affected by the noise artifacts and reveals the necessity of noise augmentation.

\noindent \textbf{Anchored Conditioning.}
Here we validate the effectiveness of anchor frame.
We manually set the anchor frame to be zero and keep the model structure unchanged.
As can be seen from \cref{fig:ablation_num} (a), without the anchor frame as an additional condition, the model shows a clear performance drop in terms of FVD for all evaluation datasets.
The visual results of \cref{fig:ablation_vis} showcase the obvious domain shifting problem with loss of high frequency details when removing anchor frame.
These results indicate the anchored conditioning is an essential to retain the overall appearance.

\section{Conclusion}
\label{sec:conclusion}

This paper realizes a novel text-to-video generation system, termed ART$\bigcdot$V, to generate videos conditioned on texts or images in an auto-regressive frame generation manner.
To address the error accumulation problem and support long video generation, we implement our ART$\bigcdot$V generation system by proposing mask diffusion model that carefully utilizes the priors of reference images, noise augmentation that closes the train-test discrepancy and anchored conditioning that assures scene consistency.
As validated by comprehensive experiments, we demonstrates superior performance over various comparison methods.

{
    \small

\begin{thebibliography}{78}
\providecommand{\natexlab}[1]{#1}
\providecommand{\url}[1]{\texttt{#1}}
\expandafter\ifx\csname urlstyle\endcsname\relax
  \providecommand{\doi}[1]{doi: #1}\else
  \providecommand{\doi}{doi: \begingroup \urlstyle{rm}\Url}\fi

\bibitem[dal()]{dalle2023}
https://openai.com/dall-e-3.

\bibitem[gen()]{gen2022}
https://research.runwayml.com/gen2.

\bibitem[i2v()]{i2vgenxl}
https://modelscope.cn/models/damo/Image-to-Video/summary.

\bibitem[mid()]{midjourney2022}
https://www.midjourney.com/home.

\bibitem[mot()]{motiondetect}
https://github.com/Breakthrough/PySceneDetect.

\bibitem[pev()]{pevels}
https://huggingface.co/datasets/Corran/pexelvideos.

\bibitem[t2i()]{t2iadaptersetting}
https://github.com/TencentARC/T2I-Adapter.

\bibitem[An et~al.(2023)An, Zhang, Yang, Gupta, Huang, Luo, and Yin]{an2023latent}
Jie An, Songyang Zhang, Harry Yang, Sonal Gupta, Jia-Bin Huang, Jiebo Luo, and Xi Yin.
\newblock Latent-shift: Latent diffusion with temporal shift for efficient text-to-video generation.
\newblock \emph{arXiv preprint arXiv:2304.08477}, 2023.

\bibitem[Bain et~al.(2021)Bain, Nagrani, Varol, and Zisserman]{bain2021frozen}
Max Bain, Arsha Nagrani, G{\"u}l Varol, and Andrew Zisserman.
\newblock Frozen in time: A joint video and image encoder for end-to-end retrieval.
\newblock In \emph{Proceedings of the IEEE/CVF International Conference on Computer Vision}, pages 1728--1738, 2021.

\bibitem[Balaji et~al.(2022)Balaji, Nah, Huang, Vahdat, Song, Kreis, Aittala, Aila, Laine, Catanzaro, et~al.]{balaji2022ediffi}
Yogesh Balaji, Seungjun Nah, Xun Huang, Arash Vahdat, Jiaming Song, Karsten Kreis, Miika Aittala, Timo Aila, Samuli Laine, Bryan Catanzaro, et~al.
\newblock ediffi: Text-to-image diffusion models with an ensemble of expert denoisers.
\newblock \emph{arXiv preprint arXiv:2211.01324}, 2022.

\bibitem[Blattmann et~al.(2023)Blattmann, Rombach, Ling, Dockhorn, Kim, Fidler, and Kreis]{blattmann2023align}
Andreas Blattmann, Robin Rombach, Huan Ling, Tim Dockhorn, Seung~Wook Kim, Sanja Fidler, and Karsten Kreis.
\newblock Align your latents: High-resolution video synthesis with latent diffusion models.
\newblock In \emph{CVPR}, pages 22563--22575, 2023.

\bibitem[Chen et~al.(2023{\natexlab{a}})Chen, Xia, He, Zhang, Cun, Yang, Xing, Liu, Chen, Wang, Weng, and Shan]{chen2023videocrafter1}
Haoxin Chen, Menghan Xia, Yingqing He, Yong Zhang, Xiaodong Cun, Shaoshu Yang, Jinbo Xing, Yaofang Liu, Qifeng Chen, Xintao Wang, Chao Weng, and Ying Shan.
\newblock Videocrafter1: Open diffusion models for high-quality video generation, 2023{\natexlab{a}}.

\bibitem[Chen et~al.(2023{\natexlab{b}})Chen, Wu, Xie, Wu, Li, Xia, Xiao, and Lin]{chen2023control}
Weifeng Chen, Jie Wu, Pan Xie, Hefeng Wu, Jiashi Li, Xin Xia, Xuefeng Xiao, and Liang Lin.
\newblock Control-a-video: Controllable text-to-video generation with diffusion models.
\newblock \emph{arXiv preprint arXiv:2305.13840}, 2023{\natexlab{b}}.

\bibitem[Chen et~al.(2023{\natexlab{c}})Chen, Wang, Zhang, Zhuang, Ma, Yu, Wang, Lin, Qiao, and Liu]{chen2023seine}
Xinyuan Chen, Yaohui Wang, Lingjun Zhang, Shaobin Zhuang, Xin Ma, Jiashuo Yu, Yali Wang, Dahua Lin, Yu Qiao, and Ziwei Liu.
\newblock Seine: Short-to-long video diffusion model for generative transition and prediction.
\newblock \emph{arXiv preprint arXiv:2310.20700}, 2023{\natexlab{c}}.

\bibitem[Esser et~al.(2023)Esser, Chiu, Atighehchian, Granskog, and Germanidis]{esser2023structure}
Patrick Esser, Johnathan Chiu, Parmida Atighehchian, Jonathan Granskog, and Anastasis Germanidis.
\newblock Structure and content-guided video synthesis with diffusion models.
\newblock In \emph{Proceedings of the IEEE/CVF International Conference on Computer Vision}, pages 7346--7356, 2023.

\bibitem[Fei et~al.(2023)Fei, Wu, Ji, Zhang, and Chua]{fei2023empowering}
Hao Fei, Shengqiong Wu, Wei Ji, Hanwang Zhang, and Tat-Seng Chua.
\newblock Empowering dynamics-aware text-to-video diffusion with large language models.
\newblock \emph{arXiv preprint arXiv:2308.13812}, 2023.

\bibitem[Ge et~al.(2023)Ge, Nah, Liu, Poon, Tao, Catanzaro, Jacobs, Huang, Liu, and Balaji]{ge2023preserve}
Songwei Ge, Seungjun Nah, Guilin Liu, Tyler Poon, Andrew Tao, Bryan Catanzaro, David Jacobs, Jia-Bin Huang, Ming-Yu Liu, and Yogesh Balaji.
\newblock Preserve your own correlation: A noise prior for video diffusion models.
\newblock In \emph{Proceedings of the IEEE/CVF International Conference on Computer Vision}, pages 22930--22941, 2023.

\bibitem[Gu et~al.(2023)Gu, Wang, Zhao, Lu, Zhang, Wu, Xu, Zhang, Jiang, and Xu]{gu2023reuse}
Jiaxi Gu, Shicong Wang, Haoyu Zhao, Tianyi Lu, Xing Zhang, Zuxuan Wu, Songcen Xu, Wei Zhang, Yu-Gang Jiang, and Hang Xu.
\newblock Reuse and diffuse: Iterative denoising for text-to-video generation.
\newblock \emph{arXiv preprint arXiv:2309.03549}, 2023.

\bibitem[Gupta et~al.(2018)Gupta, Schwenk, Farhadi, Hoiem, and Kembhavi]{gupta2018imagine}
Tanmay Gupta, Dustin Schwenk, Ali Farhadi, Derek Hoiem, and Aniruddha Kembhavi.
\newblock Imagine this! scripts to compositions to videos.
\newblock In \emph{ECCV}, pages 598--613, 2018.

\bibitem[He et~al.(2022)He, Yang, Zhang, Shan, and Chen]{he2022latent}
Yingqing He, Tianyu Yang, Yong Zhang, Ying Shan, and Qifeng Chen.
\newblock Latent video diffusion models for high-fidelity video generation with arbitrary lengths.
\newblock \emph{arXiv preprint arXiv:2211.13221}, 2022.

\bibitem[Ho and Salimans(2022)]{ho2022classifier}
Jonathan Ho and Tim Salimans.
\newblock Classifier-free diffusion guidance.
\newblock \emph{arXiv preprint arXiv:2207.12598}, 2022.

\bibitem[Ho et~al.(2020)Ho, Jain, and Abbeel]{ho2020denoising}
Jonathan Ho, Ajay Jain, and Pieter Abbeel.
\newblock Denoising diffusion probabilistic models.
\newblock \emph{NeuIPS}, 33:\penalty0 6840--6851, 2020.

\bibitem[Ho et~al.(2022{\natexlab{a}})Ho, Chan, Saharia, Whang, Gao, Gritsenko, Kingma, Poole, Norouzi, Fleet, et~al.]{ho2022imagen}
Jonathan Ho, William Chan, Chitwan Saharia, Jay Whang, Ruiqi Gao, Alexey Gritsenko, Diederik~P Kingma, Ben Poole, Mohammad Norouzi, David~J Fleet, et~al.
\newblock Imagen video: High definition video generation with diffusion models.
\newblock \emph{arXiv preprint arXiv:2210.02303}, 2022{\natexlab{a}}.

\bibitem[Ho et~al.(2022{\natexlab{b}})Ho, Salimans, Gritsenko, Chan, Norouzi, and Fleet]{ho2022video}
Jonathan Ho, Tim Salimans, Alexey Gritsenko, William Chan, Mohammad Norouzi, and David~J Fleet.
\newblock Video diffusion models.
\newblock \emph{arXiv:2204.03458}, 2022{\natexlab{b}}.

\bibitem[Hong et~al.(2022)Hong, Ding, Zheng, Liu, and Tang]{hong2022cogvideo}
Wenyi Hong, Ming Ding, Wendi Zheng, Xinghan Liu, and Jie Tang.
\newblock Cogvideo: Large-scale pretraining for text-to-video generation via transformers.
\newblock \emph{arXiv preprint arXiv:2205.15868}, 2022.

\bibitem[Huang et~al.(2022)Huang, Chia, Yu, Yee, K{\"u}ster, Krumhuber, Herremans, and Roig]{huang2022single}
Jiahui Huang, Yew~Ken Chia, Samson Yu, Kevin Yee, Dennis K{\"u}ster, Eva~G Krumhuber, Dorien Herremans, and Gemma Roig.
\newblock Single image video prediction with auto-regressive gans.
\newblock \emph{Sensors}, 22\penalty0 (9):\penalty0 3533, 2022.

\bibitem[Kalchbrenner et~al.(2017)Kalchbrenner, Oord, Simonyan, Danihelka, Vinyals, Graves, and Kavukcuoglu]{kalchbrenner2017video}
Nal Kalchbrenner, A{\"a}ron Oord, Karen Simonyan, Ivo Danihelka, Oriol Vinyals, Alex Graves, and Koray Kavukcuoglu.
\newblock Video pixel networks.
\newblock In \emph{ICML}, pages 1771--1779. PMLR, 2017.

\bibitem[Li et~al.(2023)Li, Chu, Wu, Yuan, Liu, Zhang, Li, Feng, Ding, and Wang]{li2023videogen}
Xin Li, Wenqing Chu, Ye Wu, Weihang Yuan, Fanglong Liu, Qi Zhang, Fu Li, Haocheng Feng, Errui Ding, and Jingdong Wang.
\newblock Videogen: A reference-guided latent diffusion approach for high definition text-to-video generation.
\newblock \emph{arXiv preprint arXiv:2309.00398}, 2023.

\bibitem[Li et~al.(2016)Li, Song, Cao, Tetreault, Goldberg, Jaimes, and Luo]{li2016tgif}
Yuncheng Li, Yale Song, Liangliang Cao, Joel Tetreault, Larry Goldberg, Alejandro Jaimes, and Jiebo Luo.
\newblock Tgif: A new dataset and benchmark on animated gif description.
\newblock In \emph{Proceedings of the IEEE Conference on Computer Vision and Pattern Recognition}, pages 4641--4650, 2016.

\bibitem[Li et~al.(2018)Li, Min, Shen, Carlson, and Carin]{li2018video}
Yitong Li, Martin Min, Dinghan Shen, David Carlson, and Lawrence Carin.
\newblock Video generation from text.
\newblock In \emph{AAAI}, 2018.

\bibitem[Liu et~al.(2019)Liu, Wang, Yuan, and Zhu]{liu2019cross}
Yue Liu, Xin Wang, Yitian Yuan, and Wenwu Zhu.
\newblock Cross-modal dual learning for sentence-to-video generation.
\newblock In \emph{Proceedings of the 27th ACM international conference on multimedia}, pages 1239--1247, 2019.

\bibitem[Luo et~al.(2023)Luo, Chen, Zhang, Huang, Wang, Shen, Zhao, Zhou, and Tan]{luo2023videofusion}
Zhengxiong Luo, Dayou Chen, Yingya Zhang, Yan Huang, Liang Wang, Yujun Shen, Deli Zhao, Jingren Zhou, and Tieniu Tan.
\newblock Videofusion: Decomposed diffusion models for high-quality video generation.
\newblock In \emph{CVPR}, pages 10209--10218, 2023.

\bibitem[Marwah et~al.(2017)Marwah, Mittal, and Balasubramanian]{marwah2017attentive}
Tanya Marwah, Gaurav Mittal, and Vineeth~N Balasubramanian.
\newblock Attentive semantic video generation using captions.
\newblock In \emph{ICCV}, pages 1426--1434, 2017.

\bibitem[Mittal et~al.(2017)Mittal, Marwah, and Balasubramanian]{mittal2017sync}
Gaurav Mittal, Tanya Marwah, and Vineeth~N Balasubramanian.
\newblock Sync-draw: Automatic video generation using deep recurrent attentive architectures.
\newblock In \emph{Proceedings of the 25th ACM international conference on Multimedia}, pages 1096--1104, 2017.

\bibitem[Mou et~al.(2023)Mou, Wang, Xie, Zhang, Qi, Shan, and Qie]{mou2023t2i}
Chong Mou, Xintao Wang, Liangbin Xie, Jian Zhang, Zhongang Qi, Ying Shan, and Xiaohu Qie.
\newblock T2i-adapter: Learning adapters to dig out more controllable ability for text-to-image diffusion models.
\newblock \emph{arXiv preprint arXiv:2302.08453}, 2023.

\bibitem[Nichol and Dhariwal(2021)]{nichol2021improved}
Alexander~Quinn Nichol and Prafulla Dhariwal.
\newblock Improved denoising diffusion probabilistic models.
\newblock In \emph{ICML}, pages 8162--8171. PMLR, 2021.

\bibitem[Pan et~al.(2017)Pan, Qiu, Yao, Li, and Mei]{pan2017create}
Yingwei Pan, Zhaofan Qiu, Ting Yao, Houqiang Li, and Tao Mei.
\newblock To create what you tell: Generating videos from captions.
\newblock In \emph{Proceedings of the 25th ACM international conference on Multimedia}, pages 1789--1798, 2017.

\bibitem[Parmar et~al.(2022)Parmar, Zhang, and Zhu]{parmar2022aliased}
Gaurav Parmar, Richard Zhang, and Jun-Yan Zhu.
\newblock On aliased resizing and surprising subtleties in gan evaluation.
\newblock In \emph{Proceedings of the IEEE/CVF Conference on Computer Vision and Pattern Recognition}, pages 11410--11420, 2022.

\bibitem[Paszke et~al.(2017)Paszke, Gross, Chintala, Chanan, Yang, DeVito, Lin, Desmaison, Antiga, and Lerer]{paszke2017automatic}
Adam Paszke, Sam Gross, Soumith Chintala, Gregory Chanan, Edward Yang, Zachary DeVito, Zeming Lin, Alban Desmaison, Luca Antiga, and Adam Lerer.
\newblock Automatic differentiation in pytorch.
\newblock 2017.

\bibitem[Podell et~al.(2023)Podell, English, Lacey, Blattmann, Dockhorn, M{\"u}ller, Penna, and Rombach]{podell2023sdxl}
Dustin Podell, Zion English, Kyle Lacey, Andreas Blattmann, Tim Dockhorn, Jonas M{\"u}ller, Joe Penna, and Robin Rombach.
\newblock Sdxl: Improving latent diffusion models for high-resolution image synthesis.
\newblock \emph{arXiv preprint arXiv:2307.01952}, 2023.

\bibitem[Qiu et~al.(2023)Qiu, Xia, Zhang, He, Wang, Shan, and Liu]{qiu2023freenoise}
Haonan Qiu, Menghan Xia, Yong Zhang, Yingqing He, Xintao Wang, Ying Shan, and Ziwei Liu.
\newblock Freenoise: Tuning-free longer video diffusion via noise rescheduling.
\newblock \emph{arXiv preprint arXiv:2310.15169}, 2023.

\bibitem[Radford et~al.(2021)Radford, Kim, Hallacy, Ramesh, Goh, Agarwal, Sastry, Askell, Mishkin, Clark, et~al.]{radford2021learning}
Alec Radford, Jong~Wook Kim, Chris Hallacy, Aditya Ramesh, Gabriel Goh, Sandhini Agarwal, Girish Sastry, Amanda Askell, Pamela Mishkin, Jack Clark, et~al.
\newblock Learning transferable visual models from natural language supervision.
\newblock In \emph{International conference on machine learning}, pages 8748--8763. PMLR, 2021.

\bibitem[Rakhimov et~al.(2020)Rakhimov, Volkhonskiy, Artemov, Zorin, and Burnaev]{rakhimov2020latent}
Ruslan Rakhimov, Denis Volkhonskiy, Alexey Artemov, Denis Zorin, and Evgeny Burnaev.
\newblock Latent video transformer.
\newblock \emph{arXiv preprint arXiv:2006.10704}, 2020.

\bibitem[Ramesh et~al.(2022)Ramesh, Dhariwal, Nichol, Chu, and Chen]{ramesh2022hierarchical}
Aditya Ramesh, Prafulla Dhariwal, Alex Nichol, Casey Chu, and Mark Chen.
\newblock Hierarchical text-conditional image generation with clip latents.
\newblock \emph{arXiv preprint arXiv:2204.06125}, 2022.

\bibitem[Reed et~al.(2017)Reed, Oord, Kalchbrenner, Colmenarejo, Wang, Chen, Belov, and Freitas]{reed2017parallel}
Scott Reed, A{\"a}ron Oord, Nal Kalchbrenner, Sergio~G{\'o}mez Colmenarejo, Ziyu Wang, Yutian Chen, Dan Belov, and Nando Freitas.
\newblock Parallel multiscale autoregressive density estimation.
\newblock In \emph{ICML}, pages 2912--2921. PMLR, 2017.

\bibitem[Rombach et~al.(2022)Rombach, Blattmann, Lorenz, Esser, and Ommer]{rombach2022high}
Robin Rombach, Andreas Blattmann, Dominik Lorenz, Patrick Esser, and Bj{\"o}rn Ommer.
\newblock High-resolution image synthesis with latent diffusion models.
\newblock In \emph{CVPR}, pages 10684--10695, 2022.

\bibitem[Salimans et~al.(2016)Salimans, Goodfellow, Zaremba, Cheung, Radford, and Chen]{salimans2016improved}
Tim Salimans, Ian Goodfellow, Wojciech Zaremba, Vicki Cheung, Alec Radford, and Xi Chen.
\newblock Improved techniques for training gans.
\newblock \emph{Advances in neural information processing systems}, 29, 2016.

\bibitem[Schuhmann et~al.(2021)Schuhmann, Vencu, Beaumont, Kaczmarczyk, Mullis, Katta, Coombes, Jitsev, and Komatsuzaki]{schuhmann2021laion}
Christoph Schuhmann, Richard Vencu, Romain Beaumont, Robert Kaczmarczyk, Clayton Mullis, Aarush Katta, Theo Coombes, Jenia Jitsev, and Aran Komatsuzaki.
\newblock Laion-400m: Open dataset of clip-filtered 400 million image-text pairs.
\newblock \emph{arXiv preprint arXiv:2111.02114}, 2021.

\bibitem[Schuhmann et~al.(2022)Schuhmann, Beaumont, Vencu, Gordon, Wightman, Cherti, Coombes, Katta, Mullis, Wortsman, et~al.]{schuhmann2022laion}
Christoph Schuhmann, Romain Beaumont, Richard Vencu, Cade Gordon, Ross Wightman, Mehdi Cherti, Theo Coombes, Aarush Katta, Clayton Mullis, Mitchell Wortsman, et~al.
\newblock Laion-5b: An open large-scale dataset for training next generation image-text models.
\newblock \emph{Advances in Neural Information Processing Systems}, 35:\penalty0 25278--25294, 2022.

\bibitem[Seo et~al.(2022)Seo, Lee, Liu, James, and Abbeel]{seo2022harp}
Younggyo Seo, Kimin Lee, Fangchen Liu, Stephen James, and Pieter Abbeel.
\newblock Harp: Autoregressive latent video prediction with high-fidelity image generator.
\newblock In \emph{ICIP}, pages 3943--3947. IEEE, 2022.

\bibitem[Singer et~al.(2022)Singer, Polyak, Hayes, Yin, An, Zhang, Hu, Yang, Ashual, Gafni, et~al.]{singer2022make}
Uriel Singer, Adam Polyak, Thomas Hayes, Xi Yin, Jie An, Songyang Zhang, Qiyuan Hu, Harry Yang, Oron Ashual, Oran Gafni, et~al.
\newblock Make-a-video: Text-to-video generation without text-video data.
\newblock \emph{arXiv preprint arXiv:2209.14792}, 2022.

\bibitem[Sohl-Dickstein et~al.(2015)Sohl-Dickstein, Weiss, Maheswaranathan, and Ganguli]{sohl2015deep}
Jascha Sohl-Dickstein, Eric Weiss, Niru Maheswaranathan, and Surya Ganguli.
\newblock Deep unsupervised learning using nonequilibrium thermodynamics.
\newblock In \emph{ICML}, pages 2256--2265. PMLR, 2015.

\bibitem[Song and Ermon(2019)]{song2019generative}
Yang Song and Stefano Ermon.
\newblock Generative modeling by estimating gradients of the data distribution.
\newblock \emph{NeuIPS}, 32, 2019.

\bibitem[Song et~al.(2020)Song, Sohl-Dickstein, Kingma, Kumar, Ermon, and Poole]{song2020score}
Yang Song, Jascha Sohl-Dickstein, Diederik~P Kingma, Abhishek Kumar, Stefano Ermon, and Ben Poole.
\newblock Score-based generative modeling through stochastic differential equations.
\newblock \emph{ICLR}, 2020.

\bibitem[Soomro et~al.(2012)Soomro, Zamir, and Shah]{soomro2012ucf101}
Khurram Soomro, Amir~Roshan Zamir, and Mubarak Shah.
\newblock Ucf101: A dataset of 101 human actions classes from videos in the wild.
\newblock \emph{arXiv preprint arXiv:1212.0402}, 2012.

\bibitem[Unterthiner et~al.(2018)Unterthiner, Van~Steenkiste, Kurach, Marinier, Michalski, and Gelly]{unterthiner2018towards}
Thomas Unterthiner, Sjoerd Van~Steenkiste, Karol Kurach, Raphael Marinier, Marcin Michalski, and Sylvain Gelly.
\newblock Towards accurate generative models of video: A new metric \& challenges.
\newblock \emph{arXiv preprint arXiv:1812.01717}, 2018.

\bibitem[Vaswani et~al.(2017)Vaswani, Shazeer, Parmar, Uszkoreit, Jones, Gomez, Kaiser, and Polosukhin]{vaswani2017attention}
Ashish Vaswani, Noam Shazeer, Niki Parmar, Jakob Uszkoreit, Llion Jones, Aidan~N Gomez, {\L}ukasz Kaiser, and Illia Polosukhin.
\newblock Attention is all you need.
\newblock \emph{NeuIPS}, 30, 2017.

\bibitem[Villegas et~al.(2022)Villegas, Babaeizadeh, Kindermans, Moraldo, Zhang, Saffar, Castro, Kunze, and Erhan]{villegas2022phenaki}
Ruben Villegas, Mohammad Babaeizadeh, Pieter-Jan Kindermans, Hernan Moraldo, Han Zhang, Mohammad~Taghi Saffar, Santiago Castro, Julius Kunze, and Dumitru Erhan.
\newblock Phenaki: Variable length video generation from open domain textual description.
\newblock \emph{arXiv preprint arXiv:2210.02399}, 2022.

\bibitem[Walker et~al.(2021)Walker, Razavi, and Oord]{walker2021predicting}
Jacob Walker, Ali Razavi, and A{\"a}ron van~den Oord.
\newblock Predicting video with vqvae.
\newblock \emph{arXiv preprint arXiv:2103.01950}, 2021.

\bibitem[Wang et~al.(2023{\natexlab{a}})Wang, Yuan, Chen, Zhang, Wang, and Zhang]{wang2023modelscope}
Jiuniu Wang, Hangjie Yuan, Dayou Chen, Yingya Zhang, Xiang Wang, and Shiwei Zhang.
\newblock Modelscope text-to-video technical report.
\newblock \emph{arXiv preprint arXiv:2308.06571}, 2023{\natexlab{a}}.

\bibitem[Wang et~al.(2023{\natexlab{b}})Wang, Yang, Tuo, He, Zhu, Fu, and Liu]{wang2023videofactory}
Wenjing Wang, Huan Yang, Zixi Tuo, Huiguo He, Junchen Zhu, Jianlong Fu, and Jiaying Liu.
\newblock Videofactory: Swap attention in spatiotemporal diffusions for text-to-video generation.
\newblock \emph{arXiv preprint arXiv:2305.10874}, 2023{\natexlab{b}}.

\bibitem[Wang et~al.(2019)Wang, Wu, Chen, Li, Wang, and Wang]{wang2019vatex}
Xin Wang, Jiawei Wu, Junkun Chen, Lei Li, Yuan-Fang Wang, and William~Yang Wang.
\newblock Vatex: A large-scale, high-quality multilingual dataset for video-and-language research.
\newblock In \emph{Proceedings of the IEEE/CVF International Conference on Computer Vision}, pages 4581--4591, 2019.

\bibitem[Wang et~al.(2023{\natexlab{c}})Wang, Yuan, Zhang, Chen, Wang, Zhang, Shen, Zhao, and Zhou]{wang2023videocomposer}
Xiang Wang, Hangjie Yuan, Shiwei Zhang, Dayou Chen, Jiuniu Wang, Yingya Zhang, Yujun Shen, Deli Zhao, and Jingren Zhou.
\newblock Videocomposer: Compositional video synthesis with motion controllability.
\newblock \emph{arXiv preprint arXiv:2306.02018}, 2023{\natexlab{c}}.

\bibitem[Wang et~al.(2023{\natexlab{d}})Wang, Chen, Ma, Zhou, Huang, Wang, Yang, He, Yu, Yang, et~al.]{wang2023lavie}
Yaohui Wang, Xinyuan Chen, Xin Ma, Shangchen Zhou, Ziqi Huang, Yi Wang, Ceyuan Yang, Yinan He, Jiashuo Yu, Peiqing Yang, et~al.
\newblock Lavie: High-quality video generation with cascaded latent diffusion models.
\newblock \emph{arXiv preprint arXiv:2309.15103}, 2023{\natexlab{d}}.

\bibitem[Weissenborn et~al.(2019)Weissenborn, T{\"a}ckstr{\"o}m, and Uszkoreit]{weissenborn2019scaling}
Dirk Weissenborn, Oscar T{\"a}ckstr{\"o}m, and Jakob Uszkoreit.
\newblock Scaling autoregressive video models.
\newblock \emph{arXiv preprint arXiv:1906.02634}, 2019.

\bibitem[Wu et~al.(2021)Wu, Huang, Zhang, Li, Ji, Yang, Sapiro, and Duan]{wu2021godiva}
Chenfei Wu, Lun Huang, Qianxi Zhang, Binyang Li, Lei Ji, Fan Yang, Guillermo Sapiro, and Nan Duan.
\newblock Godiva: Generating open-domain videos from natural descriptions.
\newblock \emph{arXiv preprint arXiv:2104.14806}, 2021.

\bibitem[Wu et~al.(2022)Wu, Liang, Ji, Yang, Fang, Jiang, and Duan]{wu2022nuwa}
Chenfei Wu, Jian Liang, Lei Ji, Fan Yang, Yuejian Fang, Daxin Jiang, and Nan Duan.
\newblock N{\"u}wa: Visual synthesis pre-training for neural visual world creation.
\newblock In \emph{ECCV}, pages 720--736. Springer, 2022.

\bibitem[Wu et~al.(2023)Wu, Ge, Wang, Lei, Gu, Shi, Hsu, Shan, Qie, and Shou]{wu2023tune}
Jay~Zhangjie Wu, Yixiao Ge, Xintao Wang, Stan~Weixian Lei, Yuchao Gu, Yufei Shi, Wynne Hsu, Ying Shan, Xiaohu Qie, and Mike~Zheng Shou.
\newblock Tune-a-video: One-shot tuning of image diffusion models for text-to-video generation.
\newblock In \emph{ICCV}, pages 7623--7633, 2023.

\bibitem[Xing et~al.(2023{\natexlab{a}})Xing, Xia, Liu, Zhang, Zhang, He, Liu, Chen, Cun, Wang, et~al.]{xing2023make}
Jinbo Xing, Menghan Xia, Yuxin Liu, Yuechen Zhang, Yong Zhang, Yingqing He, Hanyuan Liu, Haoxin Chen, Xiaodong Cun, Xintao Wang, et~al.
\newblock Make-your-video: Customized video generation using textual and structural guidance.
\newblock \emph{arXiv preprint arXiv:2306.00943}, 2023{\natexlab{a}}.

\bibitem[Xing et~al.(2023{\natexlab{b}})Xing, Dai, Hu, Wu, and Jiang]{xing2023simda}
Zhen Xing, Qi Dai, Han Hu, Zuxuan Wu, and Yu-Gang Jiang.
\newblock Simda: Simple diffusion adapter for efficient video generation.
\newblock \emph{arXiv preprint arXiv:2308.09710}, 2023{\natexlab{b}}.

\bibitem[Xu et~al.(2016)Xu, Mei, Yao, and Rui]{xu2016msr}
Jun Xu, Tao Mei, Ting Yao, and Yong Rui.
\newblock Msr-vtt: A large video description dataset for bridging video and language.
\newblock In \emph{Proceedings of the IEEE conference on computer vision and pattern recognition}, pages 5288--5296, 2016.

\bibitem[Xue et~al.(2022)Xue, Hang, Zeng, Sun, Liu, Yang, Fu, and Guo]{xue2022advancing}
Hongwei Xue, Tiankai Hang, Yanhong Zeng, Yuchong Sun, Bei Liu, Huan Yang, Jianlong Fu, and Baining Guo.
\newblock Advancing high-resolution video-language representation with large-scale video transcriptions.
\newblock In \emph{Proceedings of the IEEE/CVF Conference on Computer Vision and Pattern Recognition}, pages 5036--5045, 2022.

\bibitem[Yan et~al.(2021)Yan, Zhang, Abbeel, and Srinivas]{yan2021videogpt}
Wilson Yan, Yunzhi Zhang, Pieter Abbeel, and Aravind Srinivas.
\newblock Videogpt: Video generation using vq-vae and transformers.
\newblock \emph{arXiv preprint arXiv:2104.10157}, 2021.

\bibitem[Zhang et~al.(2023{\natexlab{a}})Zhang, Wu, Liu, Zhao, Ran, Gu, Gao, and Shou]{zhang2023show}
David~Junhao Zhang, Jay~Zhangjie Wu, Jia-Wei Liu, Rui Zhao, Lingmin Ran, Yuchao Gu, Difei Gao, and Mike~Zheng Shou.
\newblock Show-1: Marrying pixel and latent diffusion models for text-to-video generation.
\newblock \emph{arXiv preprint arXiv:2309.15818}, 2023{\natexlab{a}}.

\bibitem[Zhang et~al.(2023{\natexlab{b}})Zhang, Rao, and Agrawala]{zhang2023adding}
Lvmin Zhang, Anyi Rao, and Maneesh Agrawala.
\newblock Adding conditional control to text-to-image diffusion models.
\newblock In \emph{Proceedings of the IEEE/CVF International Conference on Computer Vision}, pages 3836--3847, 2023{\natexlab{b}}.

\bibitem[Zhang et~al.(2023{\natexlab{c}})Zhang, Wei, Jiang, Zhang, Zuo, and Tian]{zhang2023controlvideo}
Yabo Zhang, Yuxiang Wei, Dongsheng Jiang, Xiaopeng Zhang, Wangmeng Zuo, and Qi Tian.
\newblock Controlvideo: Training-free controllable text-to-video generation.
\newblock \emph{arXiv preprint arXiv:2305.13077}, 2023{\natexlab{c}}.

\bibitem[Zhao et~al.(2023)Zhao, Gu, Wu, Zhang, Liu, Wu, Keppo, and Shou]{zhao2023motiondirector}
Rui Zhao, Yuchao Gu, Jay~Zhangjie Wu, David~Junhao Zhang, Jiawei Liu, Weijia Wu, Jussi Keppo, and Mike~Zheng Shou.
\newblock Motiondirector: Motion customization of text-to-video diffusion models.
\newblock \emph{arXiv preprint arXiv:2310.08465}, 2023.

\bibitem[Zhou et~al.(2022)Zhou, Wang, Yan, Lv, Zhu, and Feng]{zhou2022magicvideo}
Daquan Zhou, Weimin Wang, Hanshu Yan, Weiwei Lv, Yizhe Zhu, and Jiashi Feng.
\newblock Magicvideo: Efficient video generation with latent diffusion models.
\newblock \emph{arXiv preprint arXiv:2211.11018}, 2022.

\end{thebibliography}

}


\newpage

\appendix

\section*{Appendix}

\section{Implementation Details}

We add more implementation details of network architecture, training and inference in this section.

\vspace{0.3cm}
\noindent \textbf{ART$\bigcdot$V Architecture.}
ART$\bigcdot$V is composed of two individual networks, \ie, mask prediction network $\Phi_{mask}$ and dynamic noise prediction network $\Phi_{dynamic}$, for estimating mask and dynamic noise, respectively.
Both networks utilize the same architecture except for the minor modifications of feature channel number.
We report the architecture details in \cref{tab:1}.
As can be seen, we reduce the feature channel of $\Phi_{mask}$ compared with $\Phi_{dynamic}$.
The parameter of $\Phi_{mask}$ is $51.18$ M, which is much smaller than $\Phi_{dynamic}$ of $1167.69$ M.
Because we utilize $\Phi_{mask}$ to predict the one-channel mask, which is easier compared with dynamic noise estimation of $\Phi_{dynamic}$.
The autoencoder and text encoder of ART$\bigcdot$V are elaborated in \cref{tab:2}.
We use AutoencoderKL \cite{rombach2022high} and FrozenOpenCLIPEmbedder \cite{radford2021learning} to initialize the autoencoder and text encoder of ART$\bigcdot$V.
We adopt the default settings of T2I-Adapter \cite{mou2023t2i} except for channel settings.
Please check the adapter setting in \cite{t2iadaptersetting}.

\vspace{0.3cm}
\noindent \textbf{Training.}
We report the training details in \cref{tab:3}.
We follow most default training settings as in \cite{rombach2022high} to train ART$\bigcdot$V.
Thanks to the 2D architecture of ART$\bigcdot$V, we can use a large batch size of 480 to conduct end-end training with the limited GPU resources.

\vspace{0.3cm}
\noindent \textbf{Inference.}
We use DPMPP2SAncestral Sampler \footnote{\url{https://github.com/Stability-AI/generative-models/blob/main/sgm/modules/diffusionmodules/sampling.py\#L247}} to conduct inference.
In order to save inference time, the sampling step is set as $50$.
We found that increasing sampling step can not bring a notable quality boot.
We choose $50$ to make a good speed-quality trade-off.
To amplify the effect of the conditional signals of reference frames $\boldsymbol{y}_{ref}$, global anchor frame $\boldsymbol{y}_{anchor}$ and text prompts $\boldsymbol{y}_{text}$, we adopt the classifier-free guidance \cite{ho2022classifier} for inference.
In specific, the final predicted noise can be formulated as 

\begin{equation}
\label{equ:7}
\begin{aligned}
\epsilon &= \Phi(\boldsymbol{y}_{ref}, \boldsymbol{y}_{anchor}, \boldsymbol{y}_{text}) \\
         &+ \omega_{ref}(\Phi(\boldsymbol{y}_{ref}, \boldsymbol{y}_{anchor}, \boldsymbol{y}_{text}) - \Phi(\emptyset, \boldsymbol{y}_{anchor}, \boldsymbol{y}_{text})) \\
         &+ \omega_{anc}(\Phi(\boldsymbol{y}_{ref}, \boldsymbol{y}_{anchor}, \boldsymbol{y}_{text}) - \Phi(\boldsymbol{y}_{ref}, \emptyset, \boldsymbol{y}_{text})) \\
         &+ \omega_{txt}(\Phi(\boldsymbol{y}_{ref}, \boldsymbol{y}_{anchor}, \boldsymbol{y}_{text}) - \Phi(\boldsymbol{y}_{ref}, \boldsymbol{y}_{anchor}, \emptyset)), \\
\end{aligned}
\end{equation}
where $\omega_{ref}$, $\omega_{anc}$ and $\omega_{txt}$ are the guidance scales of $\boldsymbol{y}_{ref}$, $\boldsymbol{y}_{anchor}$ and $\boldsymbol{y}_{text}$.
We set $\omega_{ref}$, $\omega_{anc}$ and $\omega_{txt}$ as $0.25$, $0.25$ and $6.5$, respectively.
The values may be changed for different samples to achieve better quality.

\begin{table}[h!]
    \setlength{\tabcolsep}{2pt}
    \renewcommand{\arraystretch}{1}
    \begin{center}
    \vspace{4mm}
    \caption{Network architecture details.
    We initialize $\Phi_{dynamic}$ using the pretrained SD-2.1 \cite{rombach2022high}. $\Phi_{mask}$ is randomly initialized.
    }
    \label{tab:1}
    \resizebox{1\linewidth}{!}{
    \begin{tabular}{lccc}
    \toprule[0.8pt]
    Setting & \vline & $\Phi_{dynamic}$ & $\Phi_{mask}$ \\
    \cline{1-1} \cline{3-4}
    input\_shape & \vline & $[4, 80, 80]$ & $[4, 80, 80]$ \\
    output\_shape & \vline & $[4, 80, 80]$ & $[1, 80, 80]$ \\
    model\_channels & \vline & $320$ & $64$ \\
    attention\_resolutions & \vline & $[4, 2, 1]$ & $[4, 2, 1]$ \\
    num\_res\_blocks & \vline & $2$ & $2$ \\
    channel\_mult & \vline & $[1, 2, 4, 4]$ & $[1, 2, 4, 4]$ \\
    num\_head\_channels  & \vline & $64$ & $32$ \\
    transformer\_depth & \vline & $1$ & $1$ \\
    context\_dim & \vline & $1024$ & $1024$ \\
    adapter\_config: & \vline &  &  \\
    \quad channels & \vline & $[320, 640, 1280, 1280]$ & $[64, 128, 256, 256]$ \\
    \quad nums\_rb & \vline & $2$ & $2$ \\
    \quad cin & \vline & $8$ & $8$ \\
    \quad ksize & \vline & $1$ & $1$ \\
    \quad sk & \vline & True & True \\
    \quad use\_conv & \vline & False & False \\
    \cline{1-1} \cline{3-4}
    params (M) & \vline & $1167.69$ & $51.18$ \\

    \bottomrule[0.8pt]
    \end{tabular}
    }
    \end{center}
\end{table}

\section{Model Efficiency Evaluation}
We evaluate the model efficiency of our ART$\bigcdot$V and ModelScope~\cite{wang2023modelscope} in this section.
All experiments are conducted in one Nvidia A100-80GB GPU.
Notably, ModelScope generates a whole video from text in a one-shot manner.
We compare the statistics for generating short video clips containing 16 frames.
Table \ref{tab:4} shows the results.
ART$\bigcdot$V requires slightly fewer FLOPs than ModelScope, while enjoying an almost 2$\times$ faster inference speed.
The GPU memory cost of ART$\bigcdot$V is much lower than that of ModelScope when performing inference at high resolution.
In addition, the training cost of ART$\bigcdot$V is significantly reduced compared to ModelScope, where the latter demands hundreds of GPUs to allow training on large batch size of 3200.

\section{Investigation of Masked Diffusion Model}
\begin{figure*}[t!]
    \centering
    \includegraphics[width=\textwidth]{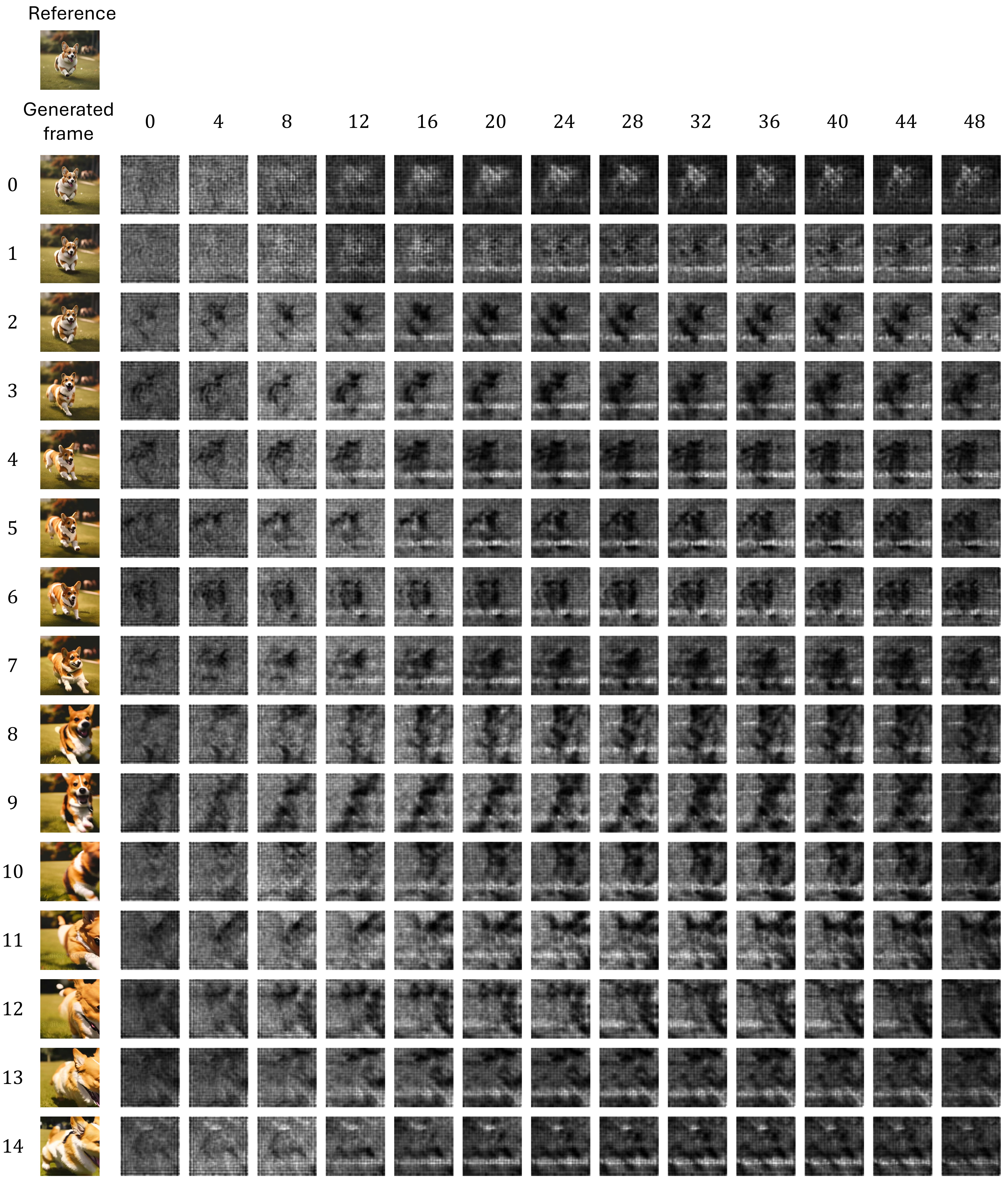}
    \caption{Visualization of mask predicted by masked diffusion model.
    }
    \label{fig:supp-mask}
\end{figure*}

In this section, we provide more visualizations of mask predicted by masked diffusion model.
The reference frame is generated by SDXL \cite{podell2023sdxl}.
The maximal sampling step is 50.
We visualize the mask with an interval of 4.
The results are presented in \cref{fig:supp-mask}.
It is worth noting that the black area of the mask has the low value, which means motion area that needs to be predicted by dynamic noise prediction network.
In contrast, the bright area of the mask has the high value, which can be directly copied from the reference frame.

\begin{table*}[t!]
    \setlength{\tabcolsep}{4pt}
    \renewcommand{\arraystretch}{1}
    \begin{center}
    \caption{Model efficiency comparisons. A batch is a video containing $16$ frames. We choose three resolution settings of $320^{2}$, $448^{2}$ and $768^{2}$ for inference. All experiments are conducted in one Nvidia A100-80GB GPU.
    }
    \vspace{-3mm}
    \label{tab:4}
    \resizebox{1\linewidth}{!}{
    \begin{tabular}{lccccccccccccccccc}
    \toprule[0.8pt]

    \multirow{4}{*}{Method} &\vline& \multicolumn{11}{c}{Inference} &\vline& \multicolumn{4}{c}{Training} \\
    \cmidrule{3-13} \cmidrule{15-18}
    &\vline& \multicolumn{3}{c}{\makecell[c]{FLOPs\\(G/batch)}} &\vline& \multicolumn{3}{c}{\makecell[c]{Throughput\\(batch/s)}} &\vline& \multicolumn{3}{c}{\makecell[c]{GPU memory\\(GB/forward)}} &\vline& \multirow{2}{*}{\makecell[c]{Batch\\size}} & \multirow{2}{*}{\makecell[c]{Iteration\\(\textit{k})}} & \multirow{2}{*}{\makecell[c]{GPU\\number}} & \multirow{2}{*}{\makecell[c]{GPU\\type}} \\
    \cline{3-5} \cline{7-9} \cline{11-13} 
    &\vline& $320^{2}$ & $448^{2}$ & $768^{2}$ &\vline& $320^{2}$ & $448^{2}$ & $768^{2}$ &\vline& $320^{2}$ & $448^{2}$ & $768^{2}$ &\vline& & & & \\
    \cline{1-1} \cline{3-5} \cline{7-9} \cline{11-13} \cline{15-18}
    ModelScope \cite{wang2023modelscope} &\vline& $3689.72$ & $7201.22$ & $21100.92$ &\vline& $7.87$ & $3.43$ & $0.78$ &\vline& $10.91$ & $16.67$ & $75.08$ &\vline& $3200$ & $267$ & - & A100-80GB \\
    ART$\bigcdot$V (Ours) &\vline& $3163.20$ & $6162.88$ & $18036.96$ &\vline& $12.16$ & $5.55$ & $1.35$ &\vline& $10.52$ & $11.08$ & $13.44$ &\vline& $480$ & $258$ &$4$ & A100-80GB \\

    \bottomrule[0.8pt]
    \end{tabular}
    }
    \end{center}
    \vspace{-4mm}
\end{table*}

\begin{table}[t!]
    \setlength{\tabcolsep}{6pt}
    \renewcommand{\arraystretch}{1}
    \begin{center}
    \caption{Details of autoencoder and text encoder. 
    }
    \vspace{-3mm}
    \label{tab:2}
    \resizebox{0.9\linewidth}{!}{
    \begin{tabular}{lcc}
    \toprule[0.8pt]
    Setting & \vline & Value \\
    \cline{1-1} \cline{3-3}
    & \vline & \multicolumn{1}{c}{\textit{\textbf{Autoencoder}}} \\
    type & \vline & AutoencoderKL \cite{rombach2022high} \\
    z\_channels & \vline & $4$ \\
    in\_channels & \vline & $3$ \\
    out\_ch & \vline & $3$ \\
    ch & \vline & $128$ \\
    ch\_mult & \vline & $[1, 2, 4, 4]$ \\
    num\_res\_blocks & \vline & $2$ \\
    & \vline & \multicolumn{1}{c}{\textit{\textbf{Textencoder}}} \\
    type & \vline & FrozenOpenCLIPEmbedder \cite{radford2021learning} \\
    Embedding dimension & \vline & 1024 \\
    CA resolutions & \vline & [1, 2, 4] \\
    CA sequence length & \vline & 77 \\

    \bottomrule[0.8pt]
    \end{tabular}
    }
    \end{center}
    \vspace{-6mm}
\end{table}

\begin{table}[t!]
    \setlength{\tabcolsep}{10pt}
    \renewcommand{\arraystretch}{1}
    \begin{center}
    \caption{Training details.}
    \vspace{-3mm}
    \label{tab:3}
    \resizebox{0.9\linewidth}{!}{
    \begin{tabular}{lcc}
    \toprule[0.8pt]
    Setting & \vline & Value \\
    \cline{1-1} \cline{3-3}
    Diffusion config: & \vline & \\
    \quad loss & \vline & mean squared error \\
    \quad timesteps & \vline & $1000$ \\
    \quad noise schedule & \vline & linear \\
    \quad linear start & \vline & $0.00085$ \\
    \quad linear end & \vline & $0.0120$ \\
    \quad prediction model & \vline & eps-pred \\
    optimizer & \vline & AdamW \\
    optimizer momentum & \vline & $\beta_1=0.9$, $\beta_2=0.999$ \\
    learning rate & \vline & $1e^{-5}$ \\
    batch size & \vline & $480$ \\
    EMA decay & \vline & $0.9999$ \\
    GPU num & \vline & 4 \\
    Training data FPS & \vline & 8 \\

    \bottomrule[0.8pt]
    \end{tabular}
    }
    \end{center}
    \vspace{-7mm}
\end{table}

\section{Additional Experiments}
\begin{figure*}[!t]
    \centering
    \includegraphics[width=1\textwidth]{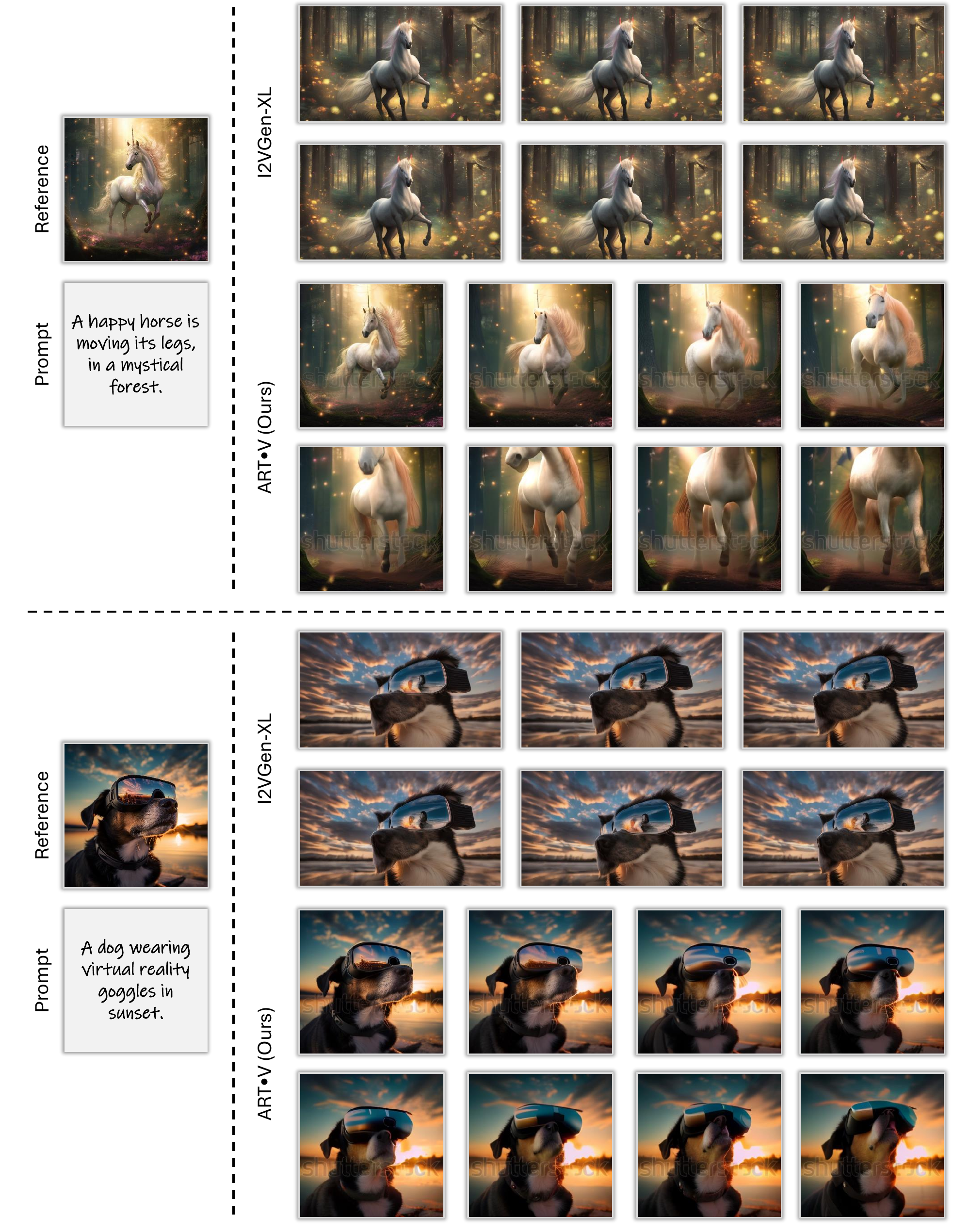}
    \caption{Visual results of text-image-to-video generation.
    }
    \label{fig:supp-ti2v-00}
\end{figure*}

\begin{figure*}[!t]
    \centering
    \includegraphics[width=1\textwidth]{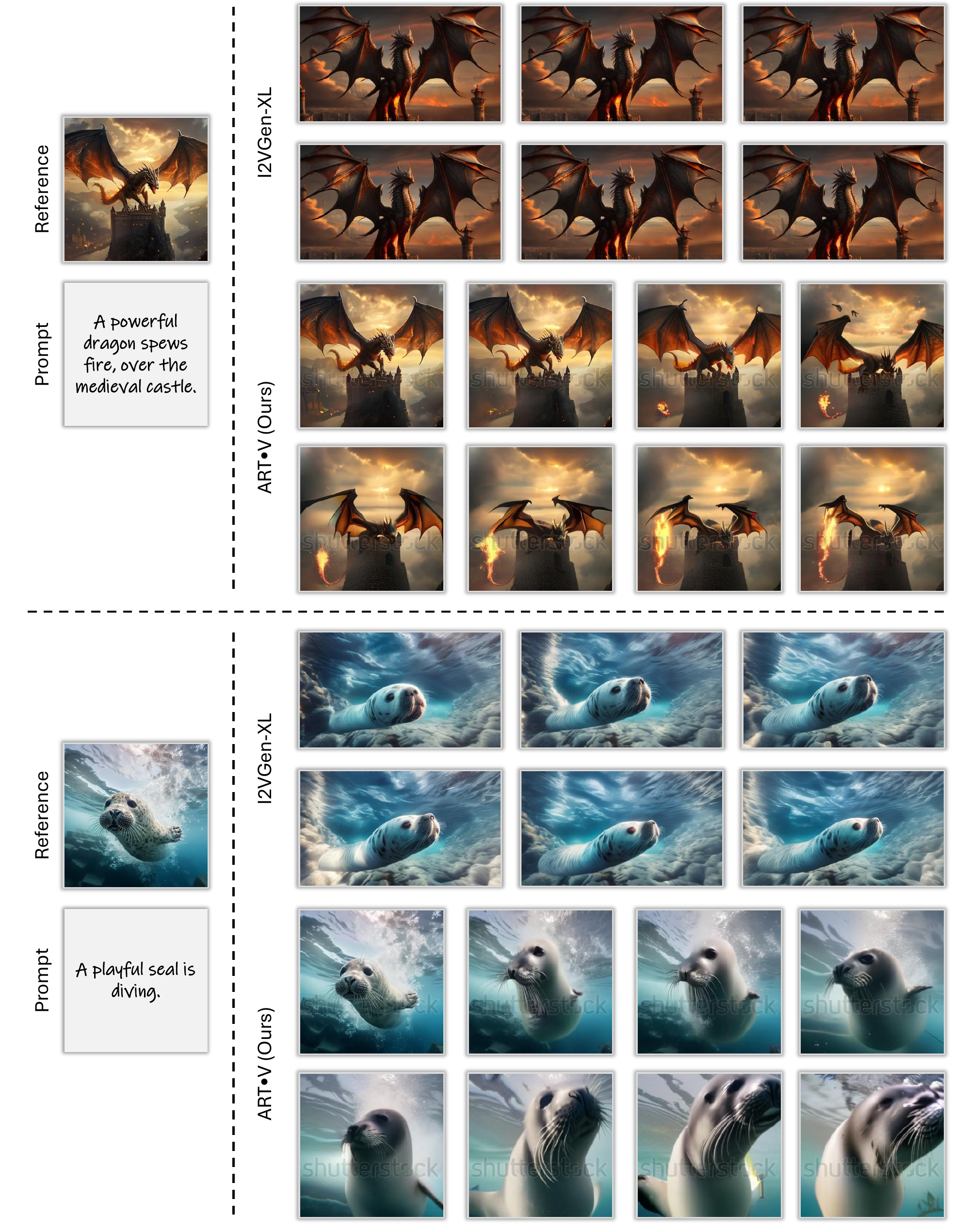}
    \caption{Visual results of text-image-to-video generation.
    }
    \label{fig:supp-ti2v-01}
\end{figure*}

\begin{figure*}[!t]
    \centering
    \includegraphics[width=1\textwidth]{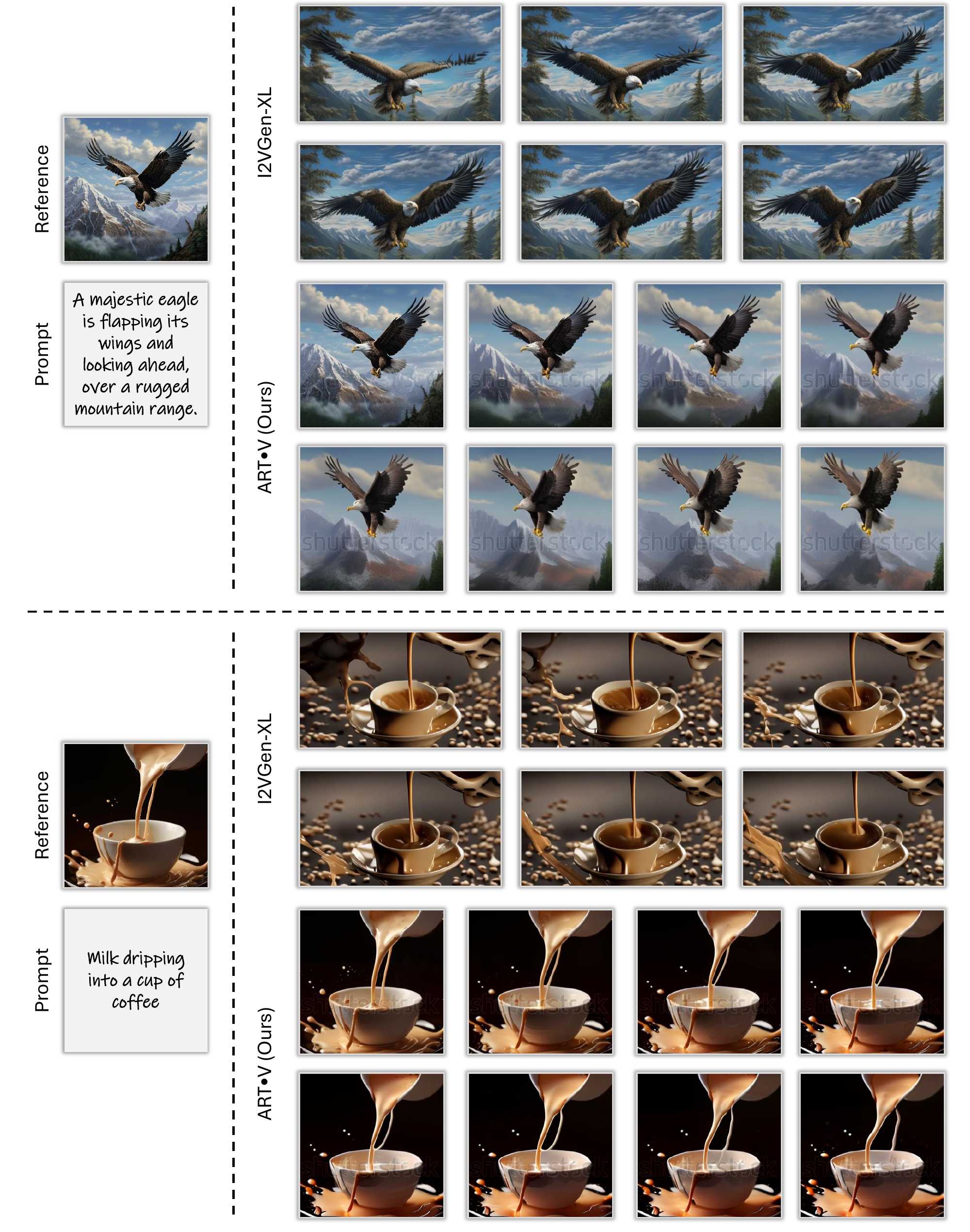}
    \caption{Visual results of text-image-to-video generation.
    }
    \label{fig:supp-ti2v-02}
\end{figure*}

\begin{figure*}[!t]
    \centering
    \includegraphics[width=1\textwidth]{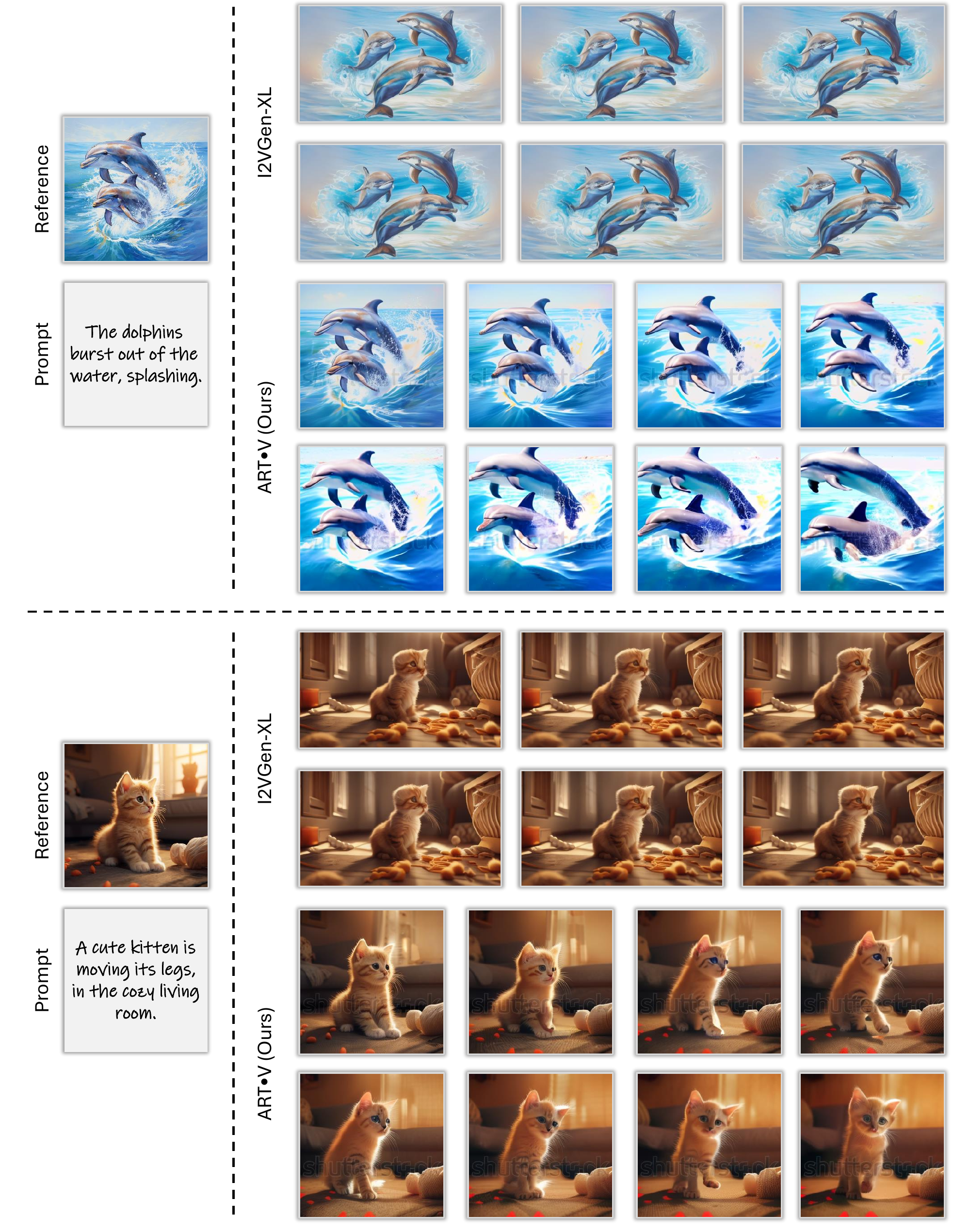}
    \caption{Visual results of text-image-to-video generation.
    }
    \label{fig:supp-ti2v-03}
\end{figure*}

\begin{figure*}[!t]
    \centering
    \includegraphics[width=1\textwidth]{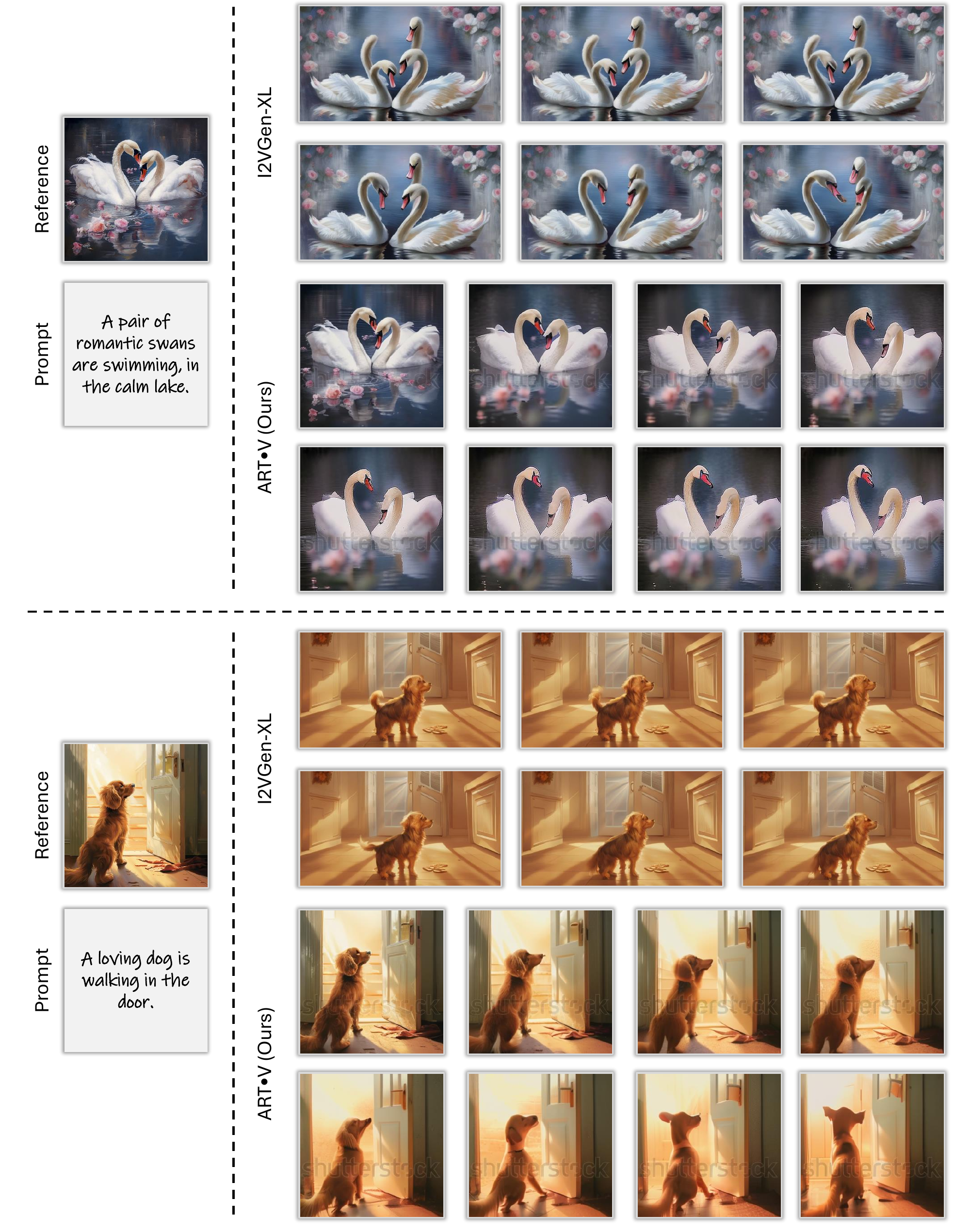}
    \caption{Visual results of text-image-to-video generation.
    }
    \label{fig:supp-ti2v-04}
\end{figure*}

\begin{figure*}[!t]
    \centering
    \includegraphics[width=0.9\textwidth]{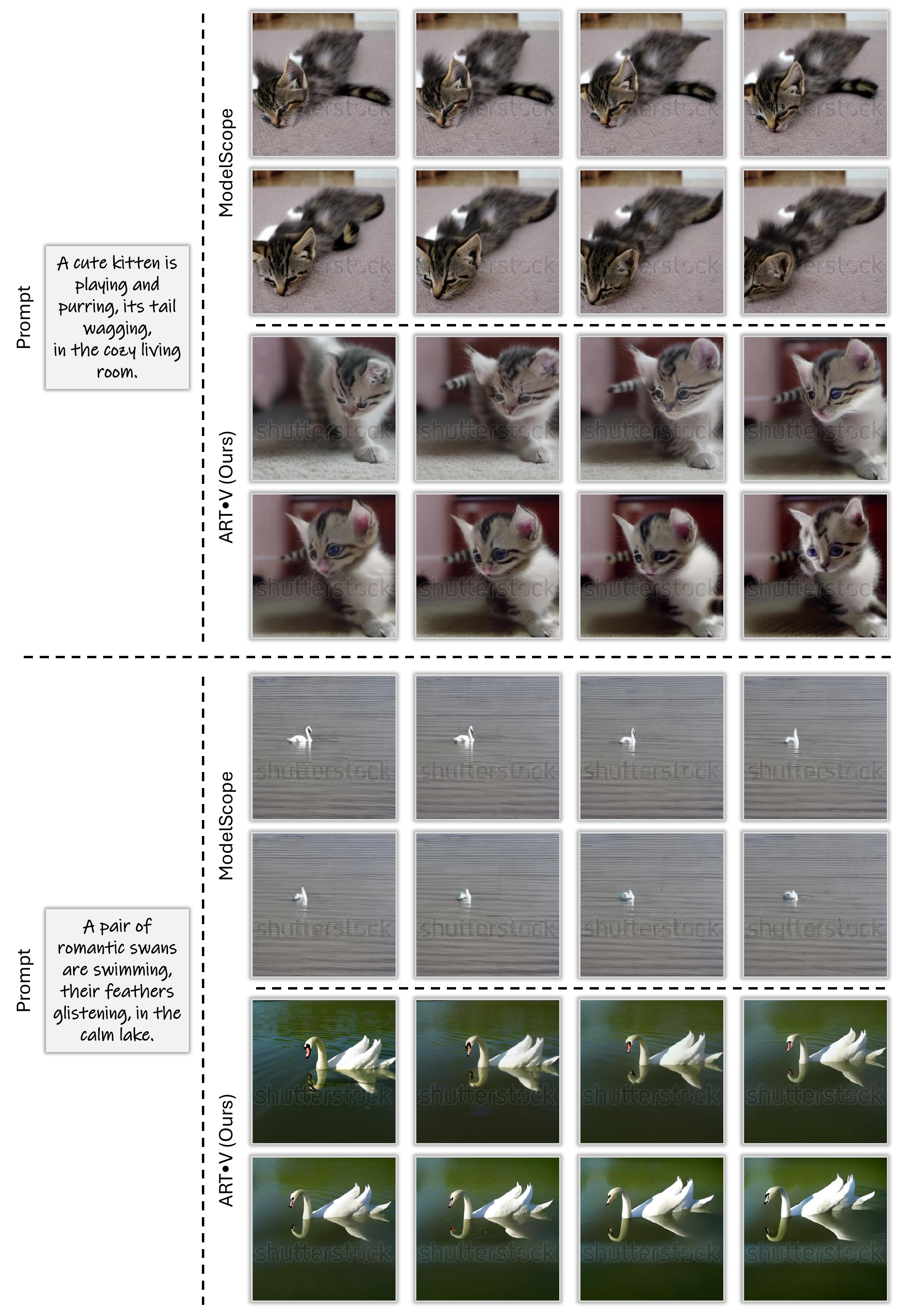}
    \vspace{-2mm}
    \caption{Visual results of text-to-video generation.
    }
    \label{fig:supp-t2v-00}
\end{figure*}

\begin{figure*}[!t]
    \centering
    \includegraphics[width=0.9\textwidth]{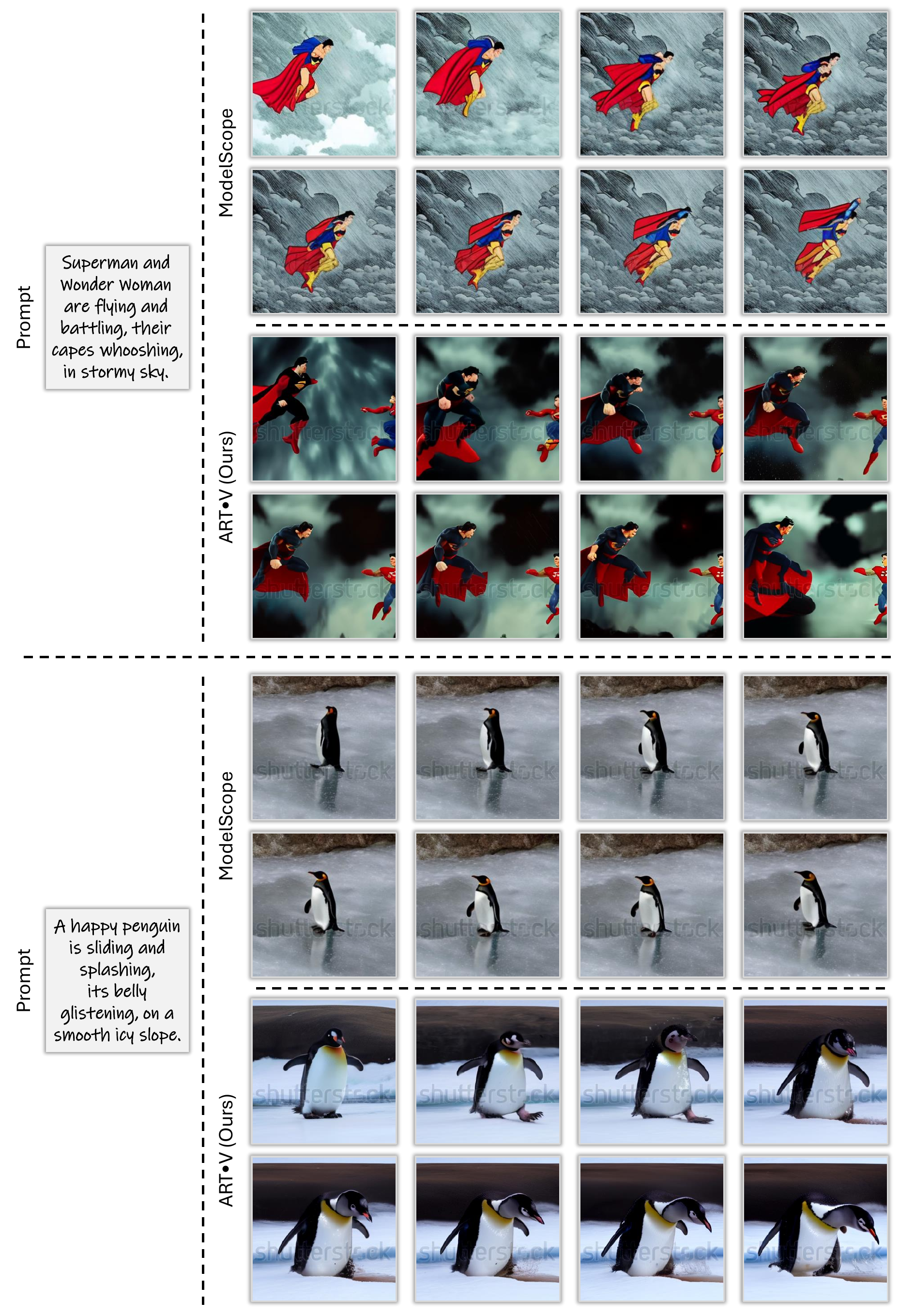}
    \vspace{-2mm}
    \caption{Visual results of text-to-video generation.
    }
    \label{fig:supp-t2v-01}
\end{figure*}

\begin{figure*}[!t]
    \centering
    \includegraphics[width=0.9\textwidth]{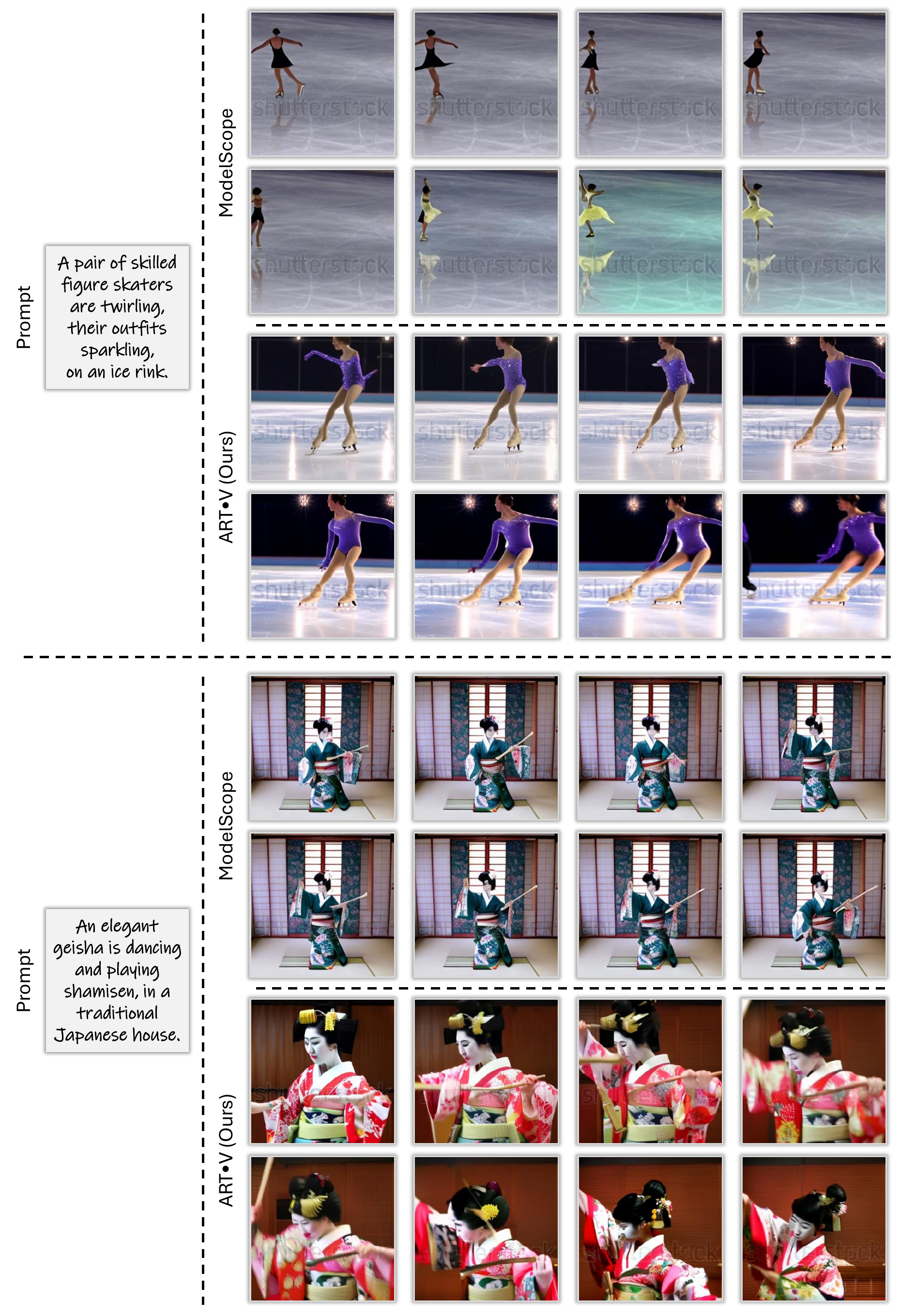}
    \vspace{-2mm}
    \caption{Visual results of text-to-video generation.
    }
    \label{fig:supp-t2v-02}
\end{figure*}

\begin{figure*}[!t]
    \centering
    \includegraphics[width=0.9\textwidth]{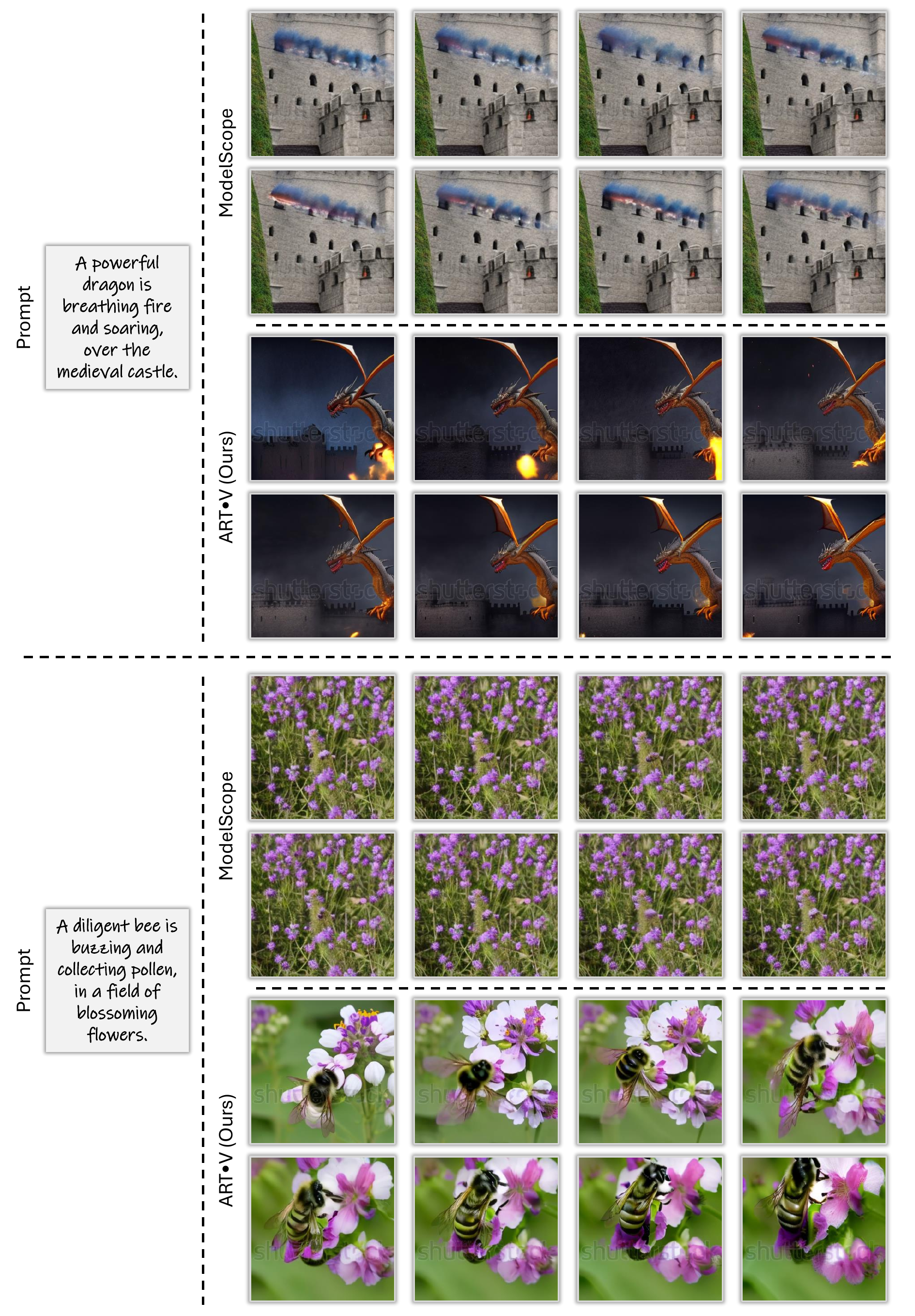}
    \vspace{-2mm}
    \caption{Visual results of text-to-video generation.
    }
    \label{fig:supp-t2v-03}
\end{figure*}

\begin{figure*}[!t]
    \centering
    \includegraphics[width=1\textwidth]{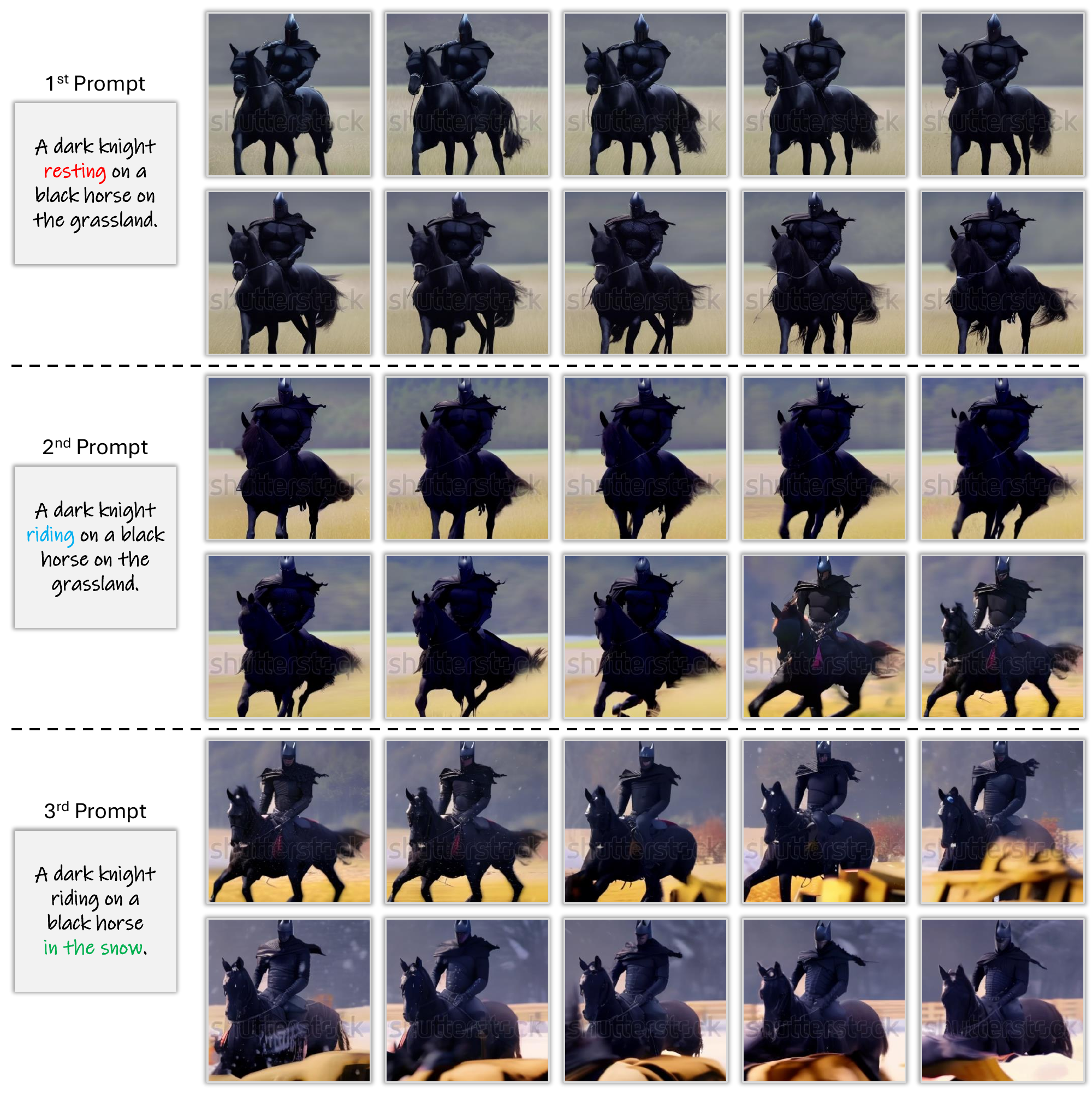}
    \caption{Visual results of multi-prompt text-to-video generation.
    }
    \label{fig:supp-mpt2v}
\end{figure*}

We provide more visual results of text-to-video generation in 
\cref{fig:supp-t2v-00}, \cref{fig:supp-t2v-01}, \cref{fig:supp-t2v-02}, \cref{fig:supp-t2v-03}, and more visual results of text-image-to-video generation in \cref{fig:supp-ti2v-00}, \cref{fig:supp-ti2v-01}, \cref{fig:supp-ti2v-02}, \cref{fig:supp-ti2v-03}, \cref{fig:supp-ti2v-04}, and more visual results of multi-prompt text-to-video generation in \cref{fig:supp-mpt2v}.

For text-to-video generation, we make comparisons with one well-known method ModelScope \cite{wang2023modelscope}.
In particular, our ART$\bigcdot$V, specifically trained for text-image-to-video generation, can generate comparable and even better results in comparison with ModelScope \cite{wang2023modelscope}. 

For text-image-to-video generation, we make comparisons with one powerful image-to-video method I2VGen-XL \cite{i2vgenxl}.
We use Midjourney \cite{midjourney2022} to generate the initial frame.
As can be clearly observed, I2VGen-XL \cite{i2vgenxl} can not keep the original details of reference frame.
It only captures the conceptual style and generate very limited and unrealistic motions.
In contrast, our ART$\bigcdot$V captures large motion,
showcasing rich details and maintaining aesthetic quality.

For multi-prompt text-to-video generation, we collect multiple prompts, each representing specified scene and motion.
We generate 16 frames for each prompt.
We fix the global anchor frame for each prompt in order to keep scene consistency.
In specific, for the first prompt, we choose the first generated frame by the model as the global anchor frame.
For the following prompts, the global anchor frame is initialized from the last generated frame of 16 frames of previous consecutive adjacent prompt.

\newpage
\FloatBarrier

\newpage
\FloatBarrier

\newpage
\FloatBarrier

\newpage
\FloatBarrier

\newpage
\FloatBarrier

\newpage
\FloatBarrier

\newpage
\FloatBarrier

\newpage
\FloatBarrier

\newpage
\FloatBarrier

\newpage
\FloatBarrier

\newpage
\FloatBarrier

\end{document}